\documentclass[a4paper,fleqn]{cas-dc}
 
\usepackage[sort,numbers]{natbib}
\usepackage[colorinlistoftodos]{todonotes}

\usepackage{ifthen}

\usepackage{fancyhdr}
\usetikzlibrary{positioning,chains}
\usetikzlibrary{shapes,arrows}
\usepackage{subcaption}

\usepackage{fancyvrb}
\usepackage{longtable}

\begin{document}
\let\WriteBookmarks\relax
\def\floatpagepagefraction{1}
\def\textpagefraction{.001}

\shorttitle{Machine Learning for Administrative Health Records}

\shortauthors{Caruana et~al.}

\title [mode = title]{Machine Learning for Administrative Health Records: A Systematic Review of Techniques and Applications}

%

%
\author[1]{Adrian Caruana}[orcid=0000-0001-7283-0220]\fnmark[1]
\cormark[1]
\credit{Conceptualization of this study, Methodology, Software}
\author[1]{Madhushi Bandara}\fnmark[1]
\author[2]{Katarzyna Musial}
\author[1,3]{Daniel Catchpoole}
\author[1,4]{Paul J. Kennedy}






\affiliation[1]{organization={Australian Artificial Intelligence Institute, Faculty of Engineering and IT, University of Technology Sydney},
   country={Australia}}
\affiliation[2]{organization={Complex Adaptive Systems Lab, Data Science Institute, Faculty of Engineering and IT, University of Technology Sydney},
	  country={Australia}}

\affiliation[3]{organization={Biospecimen Research Services, The Children's Cancer Research Unit},
   addressline={The Children's Hospital at Westmead}, 
	country={Australia}}
\affiliation[4]{organization={Joint Research Centre in AI for Health and Wellness},
	    country={University of Technology Sydney, Australia and Ontario Tech University, Canada}
}
	

%
%
%
%
%
%
%
%
\fntext[fn1]{Equal contribution.}
%

\begin{abstract}
Machine learning provides many powerful and effective techniques for analysing heterogeneous electronic health records (EHR). Administrative Health Records (AHR) are a subset of EHR collected for administrative purposes, and the use of machine learning on AHRs is a growing subfield of EHR analytics. Existing reviews of EHR analytics emphasise that the data-modality of the EHR limits the breadth of suitable machine learning techniques, and pursuable healthcare applications. Despite emphasising the importance of data modality, the literature fails to analyse which techniques and applications are relevant to AHRs. AHRs contain uniquely well-structured, categorically encoded records which are distinct from other data-modalities captured by EHRs, and they can provide valuable information pertaining to how patients interact with the healthcare system.

This paper systematically reviews AHR-based research, analysing 70\xspace relevant studies and spanning multiple databases. 
We identify and analyse which machine learning techniques are applied to AHRs and which health informatics applications are pursued in AHR-based research. We also analyse how these techniques are applied in pursuit of each application, and identify the limitations of these approaches.
We find that while AHR-based studies are disconnected from each other, the use of AHRs in health informatics research is substantial and accelerating. Our synthesis of these studies highlights the utility of AHRs for pursuing increasingly complex and diverse research objectives despite a number of pervading data- and technique-based limitations.
Finally, through our findings, we propose a set of future research directions that can enhance the utility of AHR data and machine learning techniques for health informatics research.
\end{abstract}



\maketitle

\section{Introduction}

Machine learning research is gaining huge popularity in the health informatics domain, with researchers and practitioners alike interested in utilising data-driven technologies to assist in day-to-day clinical activities. Electronic Health Records (EHR) are the clinical data repositories underpinning this research. EHRs comprise heterogeneous data modalities that can pertain to any data that is recorded in a healthcare setting. The volume of EHRs is growing rapidly, with worldwide clinical data estimated have exceeded 2 yottabytes by 2020 with an annual growth rate of 48\% \cite{pramanik2019healthcare}. Researchers, practitioners, and policymakers often use novel data analytics and machine learning techniques to understand healthcare processes, conduct clinical audits against national standards, plan future services and resource allocation, and for clinical governance \cite{shah2020secondary}.

Due to the heterogeneous nature of EHRs, the analytics techniques and healthcare applications pursued by EHR-based research depends strongly on the specific EHR data that is being analysed. For example, medical images, clinical notes, and time-series data from sensors are each common types of EHRs, yet require vastly different analytics techniques, and are useful in different aspects of healthcare research. This diversity has been established in reviews of EHR data mining applications \cite{yadav2018mining,chen2017textual}, and other surveys that review specific methods for analysing EHRs \cite{kurniati2016process,rojas2016process,erdogan2018systematic,guzzo2022process,munoz2022process,Brunson_2017,shickel2017deep,xiao2018opportunities}.

Administrative Health Records (AHR) refer specifically to the subset of EHRs that are collected for administrative purposes and comprise valuable information relating to how patients interact with the healthcare system. AHRs also have many unique characteristics that are distinct from other data-modalities captured by EHRs. AHRs contain extensive patient cohort information in the form of well-structured, categorically encoded records. In contrast, other EHR data modalities often contain unstructured or subjective records, such as clinical notes, lab reports, and medical images. 


There are three significant gaps in existing EHR-based research. Firstly, despite the importance and unique characteristics of AHRs, existing EHR-based research often does not distinguish AHRs from other EHR data-modalities, nor does the existing research examine which data-driven methods are most applicable AHRs. Secondly, the standardised, population-level context of AHRs enables insight into many unique health informatics applications that are distinct from those identified in existing surveys. Finally, studies that analyse AHRs are often disconnected from each other, making it difficult for researchers to consolidate information from other relevant studies. These gaps hinder the development of novel analytics techniques for analysing AHRs and for pursuing important population-based health informatics research questions.




The objective of this survey is to identify state-of-the-art machine learning research applied to AHRs, with a focus on studies that seek to identify population-level health service patterns. 
This study systematically analyses 70\xspace  papers selected across multiple databases to synthesise and answer four key research questions. Using the results of this review, we subsequently propose a set of future research directions that can enhance the utility of AHR data for population-level health service pattern discovery. Finally, this review seeks to add clarity to the burgeoning research area of AHR-based analytics. 

The review is organised as follows: Section~\ref{sec:background} provides background on the current research landscape of mining AHRs for health service patterns, Section~\ref{sec:methods} details the systematic literature survey methodology we followed, Section~\ref{sec:results} presents the answers to the research questions, Section~\ref{sec:openRQs} proposes some open research questions, Section~\ref{sec:discussion} discusses the findings of this review, and Section~\ref{sec:conclusion} concludes the review.


\section{Preliminaries} \label{sec:background}
\subsection{Background}

We examined several recent survey and review papers that explore the intersection of machine learning applications and AHR data to determine the scope and extent of this review. Among them, \citet{yadav2018mining} and \citet{chen2017textual} conducted comprehensive surveys on a wide range of data mining techniques used for EHRs. Other literature reviews explored the application of specific techniques to EHRs, including process mining~\cite{kurniati2016process,rojas2016process,erdogan2018systematic}, network analysis~\cite{Brunson_2017}, and deep learning \cite{shickel2017deep,xiao2018opportunities}. 

A fundamental gap in EHR-analytics literature is that AHR data is seldom distinguished from the many non-administrative data modalities comprising EHRs. Consequently, it is not clear which of the machine learning techniques identified in EHR-analytics reviews are suitable for analysing AHRs, or suitable for pursuing research objectives that pertain to AHRs. Since AHRs are used for many research objectives, (Section~\ref{sec:results:RQ2} details these objectives) this is a non-trivial gap in EHR-analytics literature.

To our knowledge, there are no existing studies that explore the broad landscape of methods suitable for analysing AHRs. This systematic review seeks to address this gap by choosing to focus on AHRs as the data source. The remainder of Section~\ref{sec:background} introduces some useful definitions to contextualise the analysis of AHRs, and outlines the scope of analytical methods that will be explored in this systematic review. 




\subsection{Definitions}\label{sec:defs}

\subsubsection{Administrative Health Records} \label{sec:ahr_char} 

According to Google Scholar, the term `electronic health records' has appeared in over one million articles, while the term `administrative health records' has only appeared in approximately one thousand articles. The use of AHRs in literature is likely much higher than the search term `administrative health records' suggests since many studies do not use this term even when the EHRs of concern are administrative. This ambiguity is problematic, since EHRs do not exclusively refer to administrative data, but encompass many other non-administrative modalities of healthcare data. 

\citet{cadarette2015introduction} provide an introduction to administrative healthcare data from a pharmacology perspective, describing five common administrative databases and variables from Ontario. We use their descriptions, as well as examples of AHRs from Australia \cite{AIHW_AHR} and the United Kingdom \cite{NHS_AHR} to inform the following definition and characterisation of AHRs.


Administrative health records are any data that is generated during interactions between various entities within a healthcare system, including but not limited to patients, physicians, hospitals, pharmacies, or government bodies. AHRs have the following five key characteristics:

\begin{itemize}
	\item AHRs principally record attributes that are \textit{discrete} or \textit{categorical} in nature. Common examples include patient attributes, diagnosis or drug codes, hospital admissions, insurance claims, or death records. 
	\item The attributes recorded in AHRs are typically \textit{temporal}, meaning that they are recordings of events that occurred at a given time. 
	\item AHRs contain linkage variables, such as patient ID, physician ID, or hospital ID, which facilitate the linkage of multiple AHRs. 
	\item AHRs are typically \textit{multivariate}, meaning that they can contain multiple attributes. For example, patient records may contain both diagnosis and treatment data.
	\item AHR attributes may also be \textit{hierarchical}. This is common of attributes that are well-defined, highly structured, or when the set of values for the attribute is large. 
\end{itemize}

\subsubsection{Population Health Service Pattern Mining} \label{sec:plhsp_char}

AHR-based research often describes the process of learning healthcare patterns  from AHRs in many different ways. In some cases, authors contextualise their work within a broad research context, such as process mining or healthcare informatics. However, most studies simply emphasise the techniques used or the objectives sought. These studies lack a shared vocabulary despite the common goal of learning about healthcare patterns from AHRs.

To clarify this, we use the term \textit{health service pattern mining} (HSPM) to describe this shared goal. This term is a combination of both \textit{health informatics} and \textit{data mining}, and describes the process of analysing patient-level event data to gain insights into the operational processes within healthcare. Section~\ref{sec:results:RQ2} details some of the applications of HSPM, and Section~\ref{sec:results:RQ3} analyses how HSPM is done in practice.

We also refer to \citet{kindig2003population} for a definition of \textit{population health}. Population health studies often use AHRs to pursue research questions concerning healthcare policies and interventions. This may be achieved by pursuing an intermediate research objective that is measurable, such as analysing health outcomes or patterns of care. An improved understanding of population health can be achieved by applying HSPM techniques to these measurable intermediate objectives.

\subsection{Technique Scope}

The scope of techniques explored in this review is informed by the definitions outlined in Section~\ref{sec:defs}. In many cases, distinct analytical techniques share similarities in the way they represent, model, or evaluate data. However, in this review we distinguish between techniques specifically by their \textit{modelling} approach. Consequently, the modelling techniques we explore may share commonalities. For example, this review specifies Bayesian networks and Markov models as distinct techniques, despite each using graphs to represent data. The remainder of this section uses this schema to outline and justify which specific analytics techniques fall within the scope of the review. 

\subsubsection{Machine Learning} \label{machine_learning}

The majority of the identified studies employ techniques that fall under the definition of \textit{machine learning}. These methods include clustering, neural networks, regression analysis, Markov models, topic modelling, ensemble learning, Bayesian networks, association rule mining, sequence mining, temporal signature mining, and dimensionality reduction. Therefore, this systematic review focusses on machine learning as the principal method for analysing AHRs. 

\subsubsection{Process Mining} \label{process_mining}

From our research, we observed that process mining was a widely explored technique for analysing EHRs, with multiple survey studies detailing its applications in healthcare process analysis \cite{kurniati2016process,rojas2016process,erdogan2018systematic,guzzo2022process,munoz2022process}. Despite widespread adoption, many studies have identified significant limitations with process mining for EHR analysis.

\citet{guzzo2022process} describe two significant limitations with this approach. Firstly, process mining struggles to accurately identify processes in EHRs since the EHR collection methods are not process-aware. Secondly, process discovery algorithms can struggle to generate interpretable process models from EHRs since they can vary significantly across patients. These studies suggest that AHR data are both less structured and more complex and diverse than the common business processes~\cite{rebuge2012business}. Most process mining algorithms are based on the assumption that the underlying processes happen in a structured fashion. \citet{rebuge2012business} suggests that this is a weak assumption in the context of human-centric clinical pathways; AHRs yield unstructured patterns and are often governed by decisions of individual patients and health practitioners, not by strict business processes. 
Furthermore, process mining is a technique more relevant to the domain of business process analysis and is based on rule-based, black-box tools rather than machine learning techniques. Therefore, we excluded process mining based studies from the survey.

\subsubsection{Network Analysis} \label{network_analysis_scope}

Unlike other techniques examined in this review, network science is not a machine learning technique, but rather a means of representing discrete data and a suite of metrics and algorithms that operate on such a representation \cite{newman2018networks}. In our research, we identified many studies that utilise techniques from network science in pursuit of HSPM. This is because network science-based models can naturally capture healthcare events interconnected by varied and diverse relationships.

In addition to the representational benefits offered by networks, metrics and algorithms from network science can also be used to understand the characteristics of AHRs. While these algorithms are often not powerful enough to capture complex health service patterns embedded within AHRs, they can often capture simple, informative, and interpretable patterns (see the \textit{Usage} section in Table~\ref{tab:network-stagesAndTechniques} and Section~\ref{subsec:network_analysis}).

For these reasons, we saw it prudent to include \textit{network analysis} as a separate technique that includes the set of metrics and algorithms which operate on networks. This technique is not to be confused with other, more complex techniques that use graph-based representations, such as Markov-based techniques, Bayesian networks, or graph neural networks.

\section{Research Method}\label{sec:methods}

Our study has been designed to answer the four research questions outlined in Section~\ref{sec:rqs}. These questions were used to inform the process of systematically reviewing relevant literature, including the bibliographic search process, inclusion and exclusion criteria, and the structure of the paper. 

\subsection{Research Questions}\label{sec:rqs}

\begin{enumerate}
    \item What are the machine learning techniques utilised in analysing health service patterns?
    \item What are the applications of health service patterns in identified studies?
    \item What are the strengths and weaknesses of specific machine learning techniques for analysing health service patterns?
    \item What are the limitations of existing machine learning techniques for health service pattern discovery?
\end{enumerate}

\subsection{Bibliographic Search Process}

To understand the research landscape, design keywords used for literature search, as well as synthesise the research questions, we followed an informal keyword-based search on \textit{Google Scholar}, \textit{IEEE Xplore}, and \textit{Springer Link}. 
This step helped the authors understand current state-of-the-art and identify existing literature reviews. Through this process, we identified 42 studies. By analysing the frequent words contained in titles, keywords, and abstracts of these studies, we came up with the following search string for the systematic literature search:

\begin{verbatim}
	('data mining' OR 'machine learning' OR 
	  'pattern mining' OR 'pattern recognition') AND 
	(patient OR treatment OR clinical OR healthcare) AND 
	(flow OR journey OR process OR pathway) AND
	('electronic health record' OR 
	  'administrative health record' OR 
	  'Administrative healthcare record' OR 
	  'electronic medical record')
\end{verbatim}

Even though AHRs represent a special subset of EHR data, in many cases existing literature does not make this distinction clear. Therefore, we include `electronic health record' and `electronic medical record' terms in the systematic literature search as to ensure that all relevant articles are included. Studies that explore EHR data that is not administrative in nature are filtered out later in the search process.

We followed the process proposed by \citet{petersen2008systematic} and \citet{harris2014write} and conducted the initial evidence search on five databases (\textit{IEEE Xplore}, \textit{ACM Digital Library}, \textit{ScienceDirect}, \textit{PubMed}, \textit{Springer Link}) for published research whose title or abstract meets the search string criteria outlined above. Publications of the \textit{Artificial Intelligence in Medicine} was also searched for matching studies and are reported under \textit{ScienceDirect} results. These primary sources were selected since they are prominent publications for computer science and machine learning as well as the healthcare domain. Finally, \textit{dblp} was also used to search for any literature that may have been missed by the six main databases and also to ensure the latest studies not indexed in initial databases are also included as our evidence. Findings were further extended through snowballing approach proposed by Wohlin~\cite{wohlin2014guidelines}.

\subsection{Filtering of Bibliographic Search Results}

\begin{figure}[htb]
	\centering
	\resizebox{\columnwidth}{!}{
\tikzset{
    mynode/.style={
        draw, rectangle, align=center, text width=5cm, font=\LARGE, inner sep=3ex},
    mynodewide/.style={
        draw, rectangle, align=center, text width=7cm, font=\LARGE, inner sep=3ex},
    mylabel/.style={
        draw, rectangle, align=center, rounded corners, font=\LARGE\bf, inner sep=2ex, 
        fill=cyan!30},
    mytitle/.style={
        draw, rectangle, align=center, rounded corners, font=\LARGE\bf, inner sep=2ex, 
        fill=magenta!30, minimum width=18cm},
    arrow/.style={
        very thick,->,>=stealth}
}

\begin{tikzpicture}[
    node distance=1.5cm,
    start chain=1 going below,
    every join/.style=arrow,
    ]
    \coordinate[on chain=1] (tc);
    \node[mynodewide, on chain=1] (n2)
        {\textbf{\# of records identified through database searching: 1939}\\IEEE Explore: 358,\\Springer Link: 1068\\ACM Digital Library: 329,\\ScienceDirect: 162,\\PubMed: 22.};
    \node[mynodewide, join, on chain=1] (n3)
        {\textbf{\# of records screened: 917}};
    \node[mynodewide, join, on chain=1] (n4)
        {\textbf{\# of records sought for retrieval: 283}};
    \node[mynodewide, join, on chain=1] (n5)
        {\textbf{\# of full-text articles accessed for eligibility: 234}};
    \node[mynodewide, join, on chain=1] (n6)
        {\textbf{\# of studies included in qualitative synthesis: \protect70\xspace }};

    \begin{scope}[start chain=going right]
        \chainin (n2);
        \node[mynodewide, join, on chain]
            {\textbf{\# of records removed before screening: 1022}\\Duplicate records removed: 7 \\Records marked as ineligible by automation tools: 1015 \\Records removed for other reasons: 0};
        \chainin (n3);
        \node[mynodewide, join, on chain]
            {\textbf{\# of records excluded: 634}};
        \chainin (n4);
        \node[mynodewide, join, on chain]
            {\textbf{\# of records sought unretrieved: 49}};
        \chainin (n5);
        \node[mynodewide, join, on chain]
            {\textbf{\# of full-text articles excluded: 164}\\Out of scope: 158,\\Insufficient detail: 3,\\Limited rigour: 3.};
    \end{scope}



    \begin{scope}[start chain=going below, yshift=-5cm, xshift=-5.5cm, node distance=.8cm]
        \node[mylabel, minimum height=8cm, on chain] {\rotatebox{90}{Identification}};
        \node[mylabel, minimum height=10cm, on chain] {\rotatebox{90}{Screening}};
        \node[mylabel, minimum height=0cm, on chain] {\rotatebox{90}{Included}};
    \end{scope}

    \node[mytitle, above right=-0.5cm and -5cm of tc, align=center, font=\Huge\bf] {PRISMA 2020 Flow Diagram};
\end{tikzpicture}
\xspace }
	\caption{Illustration of the PRISMA filtering process \cite{Pagen71} used on bibliographic search results.}\label{fig:prisma}
\end{figure}
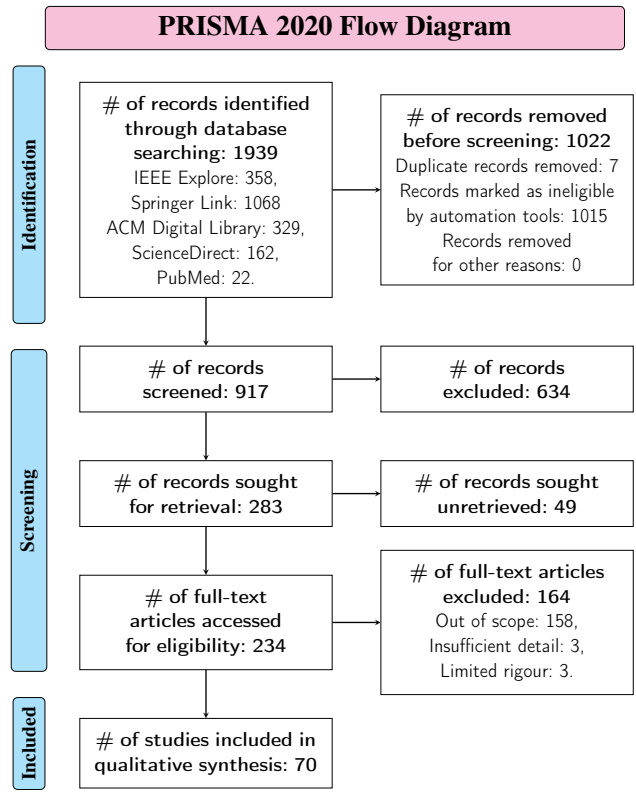

The initial database search using the keyword criteria returned many results that need to be filtered to only include works that are relevant to the four research questions outlined in Section~\ref{sec:rqs}. The filtering process is described using the PRISMA flowchart depicted in Fig.~\ref{fig:prisma}. Common examples of papers that were filtered from these results include the use of unrelated methods (e.g. natural language processing, blockchain, internet of things) or unrelated EHR modalities (e.g. images, diagnostic measurements, or unstructured text). 

Through the initial database search, we identified 1939 empirical studies as candidates for review. Among those, 917 studies were qualitatively selected as relevant studies, based on the study quality assessment and exclusion criteria. The same steps were applied to the 42 studies identified through snowballing and we identified 1 additional relevant paper for our study. One additional study was included based on expert recommendations. To avoid the inclusion of duplicate studies which would inevitably bias the result of the synthesis, we thoroughly checked if very similar studies were published in more than one paper. After assessing 234 full-texts, we eventually selected a total of 70\xspace studies to be included in the synthesis of evidence. 

\begin{figure}[h]
	\centering
	\includegraphics[width=\columnwidth]{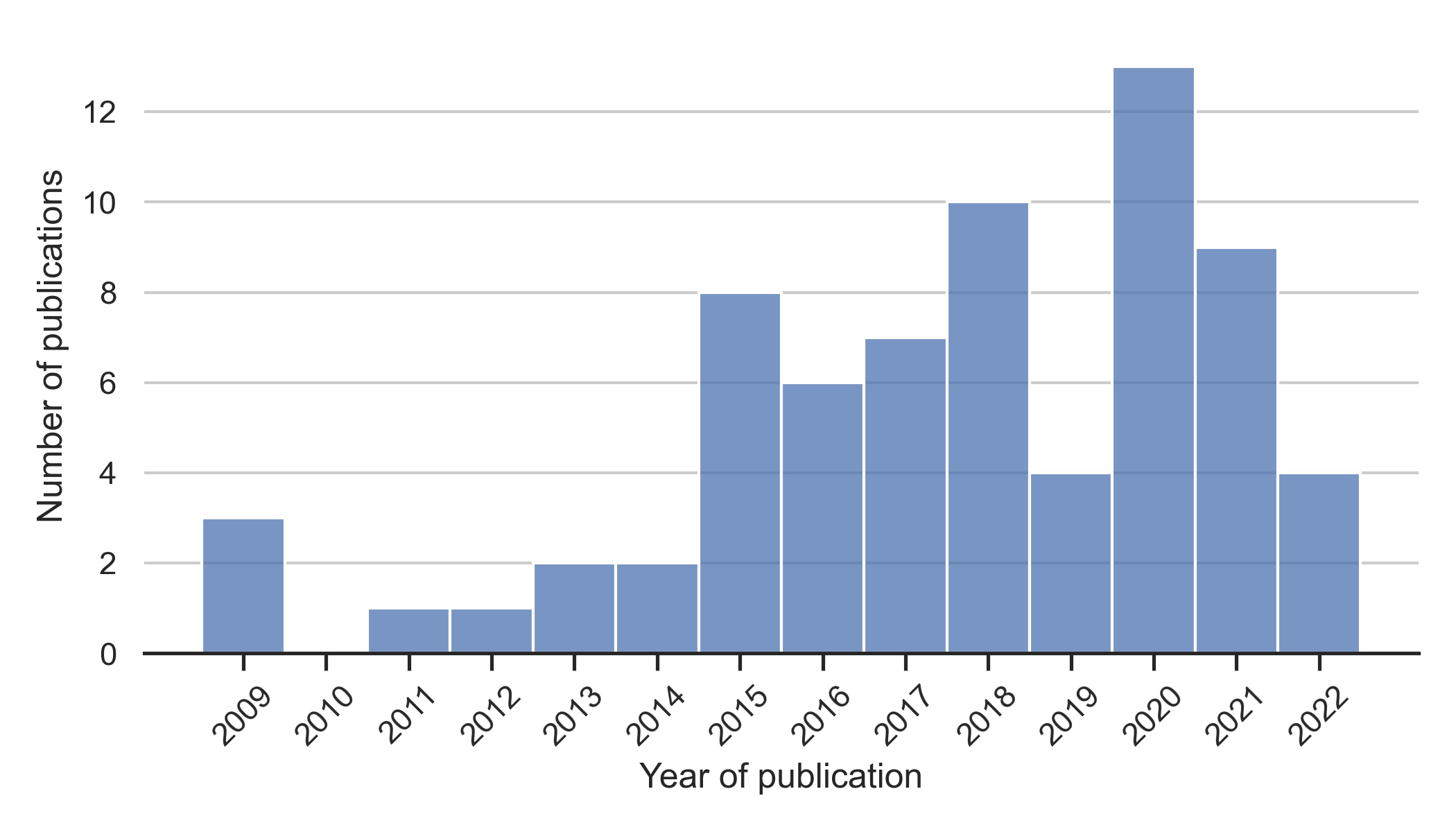}
	\caption{Histogram depicting the frequency of papers published by year for the \protect70\xspace studies used in this systematic review. This figure shows that research in this area is increasing over time, beginning around 2011. Note that year 2022 does not contain the full-year of publications.} 
	\label{fig:papers_per_year}
\end{figure}


\subsection{Exclusion Criteria and Quality Assessment}
All the solutions identified from the search phase were reviewed for relevancy. Below are the exclusion criteria we adapted from \citet{khan2011systematic} to screen studies of interest.
\begin{itemize}
	\item Books and news articles, vision papers, or front-matter
	\item Papers without any data analysis (e.g system architecture for data management, or data pre-processing methodologies)
	\item Process mining papers (see Section~\ref{process_mining} for justification)
	\item Papers that did not exclusively analyse administrative data, or whose data did not fit the characteristics of AHRs defined in Section~\ref{sec:ahr_char}
	\item Papers not written in English
	\item Papers whose full-text was not available for public access or through digital library services
	\item Papers published more than 10 years ago\footnote{Some studies published more than 10 years ago may still be included if they were selected as a part of the informal keyword search.}
\end{itemize}

To ensure the quality of the search process, the initial extraction of 42 studies and preliminary analysis were done by manual web search with a combination of different keywords, snowballing and reverse snowballing of identified papers. Two authors went through the findings and thoroughly reviewed them to define the scope of our study and search keywords for the systematic review.

Once a formal literature search was conducted, one author screened titles and abstracts of 917 records, and assessed the full-text of 234 records. In each case, the author classified the relevancy of each study as `yes', `no', and `maybe'. At least two authors reviewed the `maybe' studies and collectively decided to include or exclude the study from the evidence. Once the evidence was finalised, one author extracted results from the identified literature and a second author reviewed and verified the results to ensure accuracy.

\section{Results}\label{sec:results}

This section first presents an overview (Section~\ref{sec:results:overview}) of the 70\xspace studies identified through the bibliographic search process, before proceeding to use these studies to answer the four research questions (Sections~\ref{sec:results:RQ1}-\ref{sec:results:RQ4}).

\subsection{Overview of AHR-based Studies}\label{sec:results:overview}

Fig.~\ref{fig:papers_per_year} depicts the year of publication for each of the AHR-based studies, and shows that AHR-based research is increasing over time. Among these studies, the following were the most cited by other AHR-based studies: 

\begin{itemize}
	\item With nine citations: \cite{choi2016multi} explores methods for representation learning of medical and diagnosis codes
	\item With eight citations: \cite{huang2015mining} uses a novel topic-modelling approach to mine latent treatment patterns in AHRs.
	\item With five citations: \cite{jensen2014temporal} identifies temporal disease trajectories using a population-scale AHR database.
	\item With four citations each: \cite{liu2015temporal} uses a graph-based framework for temporal phenotyping, \cite{baker2017process} defines clinical pathways observed in chemotherapy, and \cite{choi2017gram} explores the use of healthcare representation learning using graph-based attention models. 
\end{itemize}

\begin{figure}
	\centering
	\includegraphics[width=\columnwidth]{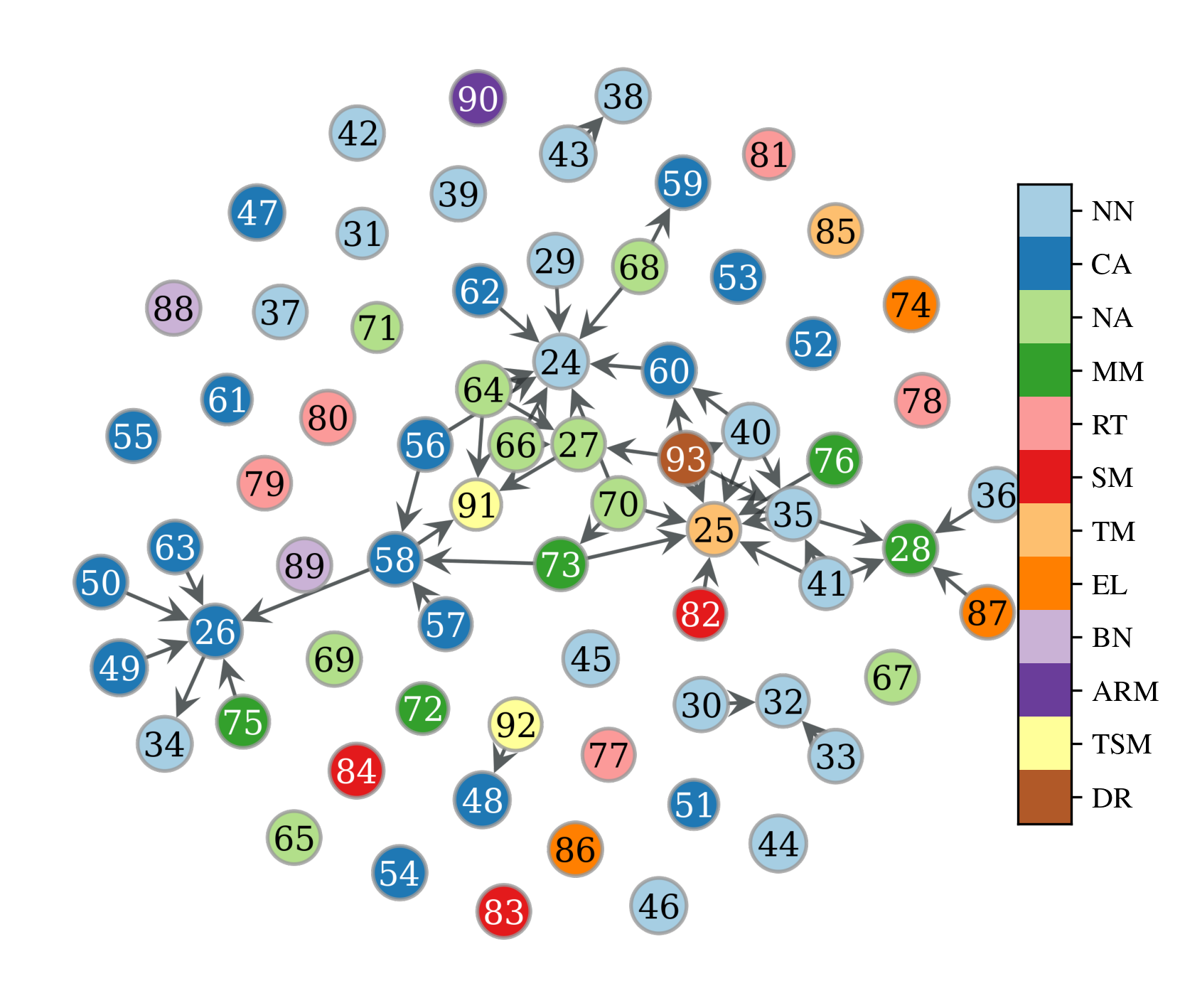}
	\caption{This citation network of the \protect70\xspace AHR-based studies reviewed in this paper. Each node in the network is one study (numbers refer to references as per the bibliography), with directed edges indicating a \textit{from}$\to$\textit{to} citation, and with node colour indicating the ML technique used in the study (as per Table~\ref{tab:ML-techniques}).} 
	\label{fig:referential_community}  
\end{figure}

Fig.~\ref{fig:referential_community} depicts the citation network of the AHR-based studies, with each study coloured by the ML technique used (Section~\ref{sec:results:RQ1} explores these ML techniques in more detail). The mean number of AHR-based citations per study is $0.8$, with a majority (53\%) of studies containing no citations to other AHR-based literature. Furthermore, while many neural network-based studies cite each other, there is no clear technique-based citation patterns for other methods used. These findings indicate that AHR-based research is highly disconnected. 

\subsection{RQ1: What are the machine learning techniques utilised in analysing health service patterns?}\label{sec:results:RQ1}

This section explores which machine learning techniques are used to analyse AHRs. We provide a detailed analysis for each of the most common techniques, and a shorter analysis for some of the infrequently used techniques. The detailed analysis includes a flow-chart of the steps followed by the studies in applying each technique, a table for each technique\footnote{Evidence only report best performing models in identified studies and do not report models developed for comparison such as baseline methods.} describing how each of the studies implement these analytical steps, and a discussion that outlines any other interesting findings from the studies. 

In total, we observed twelve types of ML techniques used in the identified studies as summarised in Table~\ref{tab:ML-techniques}. Sections \ref{sec:neuralnetworks} to \ref{sec:linearregression} include a detailed analysis of the five most common techniques, and Section \ref{sec:othertechniques} provides shorter analysis of each of the remaining techniques.

\begin{table*}[width=.9\textwidth,cols=2,pos=h!]
	\caption{AHR analysis techniques }\label{tab:ML-techniques}
	\begin{tabular}{ll}
\toprule
                       Technique &                                                                                                                                                                                                                                                                                                                           Studies \\
\midrule
            Neural Networks (NN) &                                           \cite{choi2016multi,choi2017gram,choi2018mime,guo2020comparative,pham2017predicting,zhang2018patient2vec,jin2018treatment,li2020ccae,xu2018learning,beaulieu2018mapping,Hong_2017,Li_2020,Lu_2021,Wolff_2020,steinberg2021language,Zheng_2021,Caruana_2022,Gerrard_2022,ochoa2022graph} \\
           Cluster Analysis (CA) & \cite{doshi-velez2014comorbidity,Zhang_2015,roque2011using,sideris2016flexible,chen2009cancer,jensen2014temporal,chen2018data,apunike2020analyses,johns2020clustering,bose2009trace,prokofyeva2019application,chen2019mining,aspland2021modified,bean2017network,hompes2015discovering,Chambard_2021,Mohan_Kumar_2020,Huang_2015} \\
           Network Analysis (NA) &                                                                                   \cite{roque2011using,sideris2016flexible,chen2009cancer,steinhaeuser2009network,hanauer2013modeling,glicksberg2016comparative,kannan2016conditional,bean2017network,liu2015temporal,Dong_2021,Kushima_2019,Zhang_2017,Wang_2021,ochoa2022graph} \\
               Markov Model (MM) &                                                                                                                                                                                                          \cite{Zhang_2015,maass2020markov,baker2017process,huang2018probabilistic, leontjeva2016complex,Nagrecha_2017,Bueno_2018} \\
Regression-based Techniques (RT) &                                                                                                                                                                                                                   \cite{glicksberg2016comparative,roder2021female,te2019alignment,shahabi2020differences,Li_2016,sun2022applying} \\
            Sequence Mining (SM) &                                                                                                                                                                                                                                                       \cite{Dong_2021,Kushima_2019,kaur2020time,Estiri_2020,Vincent_Paulraj_2021} \\
             Topic Modeling (TM) &                                                                                                                                                                                                                                                              \cite{huang2015mining,prokofyeva2019application,Li_2020,Huang_2015a} \\
          Ensemble Learning (EL) &                                                                                                                                                                                                                                                                                \cite{leontjeva2016complex,Boland_2015,Maali_2018} \\
          Bayesian Networks (BN) &                                                                                                                                                                                                                                                                             \cite{Nagrecha_2017,wang2020survivability,Weiss_2013} \\
   Association Rule Mining (ARM) &                                                                                                                                                                                                                                                                                        \cite{du2019variance,Vincent_Paulraj_2021} \\
 Temporal Signature Mining (TSM) &                                                                                                                                                                                                                                                                                              \cite{wang2012framework,Nguyen_2015} \\
   Dimensionality Reduction (DR) &                                                                                                                                                                                                                                                                                                             \cite{chen2020fusion} \\
\bottomrule
\end{tabular}
\xspace 
\end{table*}

\subsubsection{Neural Networks}\label{sec:neuralnetworks}

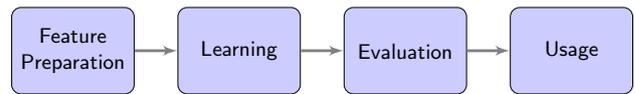
\begin{figure}[h!]
	\centering
	\tikzstyle{decision} = [diamond, draw, fill=blue!20,
    text width=4.5em, text badly centered, node distance=2.5cm, inner sep=0pt]
	\tikzstyle{block} = [rectangle, draw, fill=blue!20,
		text width=5em, text centered, rounded corners, minimum height=4em]
	\tikzstyle{line} = [draw, very thick, color=black!50, -latex']
	\tikzstyle{cloud} = [draw, ellipse,fill=red!20, node distance=2.5cm,
		minimum height=2em]
	\resizebox{\columnwidth}{!}{
		\begin{tikzpicture}[scale=1, node distance = 2.5cm, auto]
			\node [block] (init) {Feature Preparation};
			\node [block, right of=init] (Learning) {Learning};
			\node [block, right of=Learning] (Evaluation) {Evaluation};
			\node [block, right of=Evaluation] (Usage) {Usage};
			\path [line] (init) -- (Learning);
			\path [line] (Learning) -- (Evaluation);
			\path [line] (Evaluation) -- (Usage);
		\end{tikzpicture}
	}
	\caption{Steps followed in neural network-based studies}
	\label{fig:neuralNet}
\end{figure}

Neural networks are the most widely adopted machine learning technique with nineteen\xspace studies. The analysis in these studies follows the four analysis steps as in Fig.~\ref{fig:neuralNet}. Specific techniques and details associated with each step are listed in Table~\ref{tab:neuralNet-stagesAndTechniques}, with associated studies for each technique. 

When analysing the \textit{Class of Techniques} row of Table~\ref{tab:neuralNet-stagesAndTechniques}, it is clear that the main contribution proposed in neural network-based studies is a representation learning strategy to capture and summarise qualities of AHR data and to reduce dimensions of AHRs. Five studies propose end-to-end frameworks that encapsulate both representation learning and predictive learning.

In the feature preparation step, different features or combinations of features from EHR records are used with diagnosis and treatment codes used more prominently than others. Two studies go beyond common feature representation and incorporate domain knowledge to increase representation quality. In \citet{li2020ccae} and \citet{ochoa2022graph}, instead of using medical records directly, vectorised patient representations resulting from auto-encoders were used as the feature for sequence embedding and graph embedding tasks respectively.

Under the learning step, different neural network-based models are used with sequence-based based frameworks (e.g., LSTM, RNN, Transformer), auto-encoders, and sequence embedding being the most commonly applied approaches. Many studies propose novel techniques or frameworks for feature representation as well.

As most studies focus on representation learning, the main usage of the neural network model in these studies is for some downstream analytics tasks such as prediction or clustering. Two studies each use the dimensionality reduction provided by representation learning to conduct visual analytics and to generate clinically meaningful interpretations of large datasets. All ten studies evaluate their proposed solution by comparing performance, usually by conducting one or more prediction tasks and comparing performance with baseline methods for representation learning and predictive analytics.

\begin{table*}[]
	\caption{Neural Network techniques}
	\label{tab:neuralNet-stagesAndTechniques}
	\begin{tabular}{|p{0.15\textwidth}|p{0.45\textwidth}|p{0.3\textwidth}|}
		\hline
		\textbf{Step} & \textbf{Technique} & \textbf{Studies}
		\\ \hline
		
		\multirow{3}{*}
		{Class of Techniques}
		& 
		Representation Learning &	
		\cite{choi2016multi,choi2017gram,choi2018mime,guo2020comparative,pham2017predicting,zhang2018patient2vec,jin2018treatment,li2020ccae,xu2018learning,beaulieu2018mapping} \cite{Hong_2017,Lu_2021,steinberg2021language,Caruana_2022}
		\\ \cline{2-3}  &
		Predictive Learning & 
		\cite{pham2017predicting,jin2018treatment,Li_2020,Zheng_2021,Gerrard_2022,ochoa2022graph}	
		\\ \hline
		
		\multirow{6}{*}
		{Features Used}
		& 
		Diagnosis Codes &	
		\cite{choi2016multi,choi2017gram,choi2018mime,zhang2018patient2vec,jin2018treatment,Hong_2017,Li_2020,Lu_2021,steinberg2021language,Zheng_2021,Caruana_2022,Gerrard_2022,ochoa2022graph}
		\\ \cline{2-3}  &
		Treatment Information (procedure and medication codes) & 
		\cite{choi2016multi,choi2018mime,zhang2018patient2vec,jin2018treatment,li2020ccae,Hong_2017,Li_2020,Lu_2021,steinberg2021language,Zheng_2021}	
		\\ \cline{2-3}  &
		Encounter Records & 
		\cite{choi2018mime,guo2020comparative,pham2017predicting}	
		\\ \cline{2-3}  &
		Domain Knowledge (grouping) & 
		\cite{choi2017gram,choi2018mime}	
		\\ \cline{2-3}  &
		Demographic Information & 
		\cite{jin2018treatment,li2020ccae,ochoa2022graph}
		\\ \cline{2-3}  &
		Vectorised Patient Representation & 
		\cite{li2020ccae}
		\\ \hline

		\multirow{15}{*}
		{Learning Technique}
		& 
		LSTM based framework &
		\cite{pham2017predicting,jin2018treatment,beaulieu2018mapping}
		\\ \cline{2-3}  &
		Autoencoder & 
		\cite{guo2020comparative,beaulieu2018mapping}		
		\\ \cline{2-3}  &
		Sequence Embedding & 
		\cite{guo2020comparative,li2020ccae,Hong_2017}	
		\\ \cline{2-3}  &

		RNN based framework & \cite{li2020ccae}
		\\ \cline{2-3}  &
		Reinforcement learning & \cite{Zheng_2021}
		\\ \cline{2-3}  &
		Graph  Neural Network & \cite{ochoa2022graph}
		\\ \cline{2-3}  &
		Multilevel Embedding (Novel) &	\cite{choi2018mime}
		\\ \cline{2-3}  &
		GRaph-based Attention Model (novel) & \cite{choi2017gram}
		\\ \cline{2-3}  &
		Med2Vec (novel) & \cite{choi2016multi}
		\\ \cline{2-3}  &
		Patient2Vec (Novel) & 
		\cite{zhang2018patient2vec} 
		\\ \cline{2-3}  &
		RoMCP (novel) &  
		\cite{xu2018learning} 
		\\ \cline{2-3} &
		Neural Topic Model (Novel) & 
		\cite{Li_2020}
		\\ \cline{2-3}  &
		ProAID (Novel) & 
		\cite{Lu_2021} 
		\\ \cline{2-3}  &
		Clinical Language Model-based Representation (Novel) & 
		\cite{steinberg2021language}
		\\ \cline{2-3} &
		Categorical Sequence Encoder (CaSE) (Novel) & 
		\cite{Caruana_2022}
		\\ \cline{2-3} &
		Multi-task Transformer (TransMT) (Novel) & 
		\cite{Gerrard_2022}
		\\ \hline

		\multirow{4}{*}
		{Usage}
		& 
		Feature for downstream analytics & \cite{choi2016multi,choi2017gram,choi2018mime,pham2017predicting,zhang2018patient2vec,jin2018treatment,li2020ccae,xu2018learning,beaulieu2018mapping,Hong_2017} \cite{Li_2020,Lu_2021,steinberg2021language,Caruana_2022}
		\\ \cline{2-3}  &
		Visual Analytics & \cite{guo2020comparative,zhang2018patient2vec} 
		\\ \cline{2-3}  &
		Clinically meaningful interpretations & \cite{choi2016multi,zhang2018patient2vec,Zheng_2021,Gerrard_2022,ochoa2022graph} 
		\\ \hline

		\multirow{2}{*}
		{Evaluation}
		& 
		Compare performance with baseline methods &	\cite{choi2016multi,choi2017gram,choi2018mime,guo2020comparative,pham2017predicting,zhang2018patient2vec,jin2018treatment,li2020ccae,xu2018learning,beaulieu2018mapping} \cite{Hong_2017,Li_2020,Lu_2021,steinberg2021language,Zheng_2021,Caruana_2022,Gerrard_2022,ochoa2022graph}
		\\ \cline{2-3} &
		Visualise and inspect clusters & 
		\cite{ochoa2022graph}
				\\ \hline
		\end{tabular}
\end{table*}
\xspace 

\subsubsection{Cluster Analysis}\label{sec:clustering}

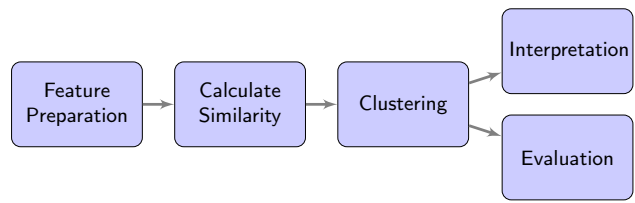
\begin{figure}[h!]
	\centering
	\tikzstyle{decision} = [diamond, draw, fill=blue!20,
    text width=4.5em, text badly centered, node distance=2.5cm, inner sep=0pt]
	\tikzstyle{block} = [rectangle, draw, fill=blue!20,
		text width=5.5em, text centered, rounded corners, minimum height=4em]
	\tikzstyle{line} = [draw, very thick, color=black!50, -latex']
	\tikzstyle{cloud} = [draw, ellipse,fill=red!20, node distance=2.5cm,
		minimum height=2em]
	\resizebox{\columnwidth}{!}{
		\begin{tikzpicture}[scale=1, node distance = 2.5cm, auto]
			\node [block] (init) {Feature Preparation};
			\node [block, right of=init] (Similarity) {Calculate Similarity};
			\node [block, right of=Similarity] (Clustering) {Clustering};
			\node [block, above right=-0.5cm and 0.5cm of Clustering] (Interpretation) {Interpretation};
			\node [block, below right=-0.5cm and 0.5cm of Clustering] (Evaluation) {Evaluation};
			\path [line] (init) -- (Similarity);
			\path [line] (Similarity) -- (Clustering);
			\path [line] (Clustering) -- (Interpretation);
			\path [line] (Clustering) -- (Evaluation);
		\end{tikzpicture}
	}

	\caption{Steps followed in cluster analysis-based studies}
	\label{fig:clustering}
\end{figure}

Cluster analysis is another widely used technique with eighteen\xspace studies utilising multiple different similarity measures and clustering algorithms. Fig. \ref{fig:clustering} illustrates the generic process steps followed in these studies. We observed that the first three steps (data representation, calculating similarity, and clustering) are followed in all studies, but some studies only followed one of the interpretation and evaluation steps.

Table \ref{tab:clustering-stagesAndTechniques} catalogues the different ways cluster analysis is used in the identified studies. 

The most widely used data representation technique is to represent patient clinical pathways as strings with different events encoded as letters. In addition to that, a variety of vector representations are used in different clustering algorithms for different purposes, including the novel model by \citet{chen2019mining} that represents the drug use duration statistics of a patient as a vector. Finally, two studies~\cite{chen2009cancer,bean2017network} utilised networks to represent the data.

To calculate similarity between events in AHRs for cluster analysis, most studies use common distance measures such as Euclidean distance, Jaccard score and Levenshtein distance, but the choice of metric depended on the data representation and the study objective. We also identified five studies that have proposed novel similarity measures (or variations on the most common measures) that can capture unique properties of health service patterns.

Multiple different clustering algorithms were utilised under the clustering step. Ward's method for hierarchical clustering was the most commonly used algorithm. Hierarchical-based clustering algorithms were also common, particularly because of the flexibility it offers to adjust the number and size of resulting clusters. 

Interpretation of clustering results was commonly achieved using descriptive analytics that incorporates visualisation techniques including heat-maps, dendrograms, and charts of statistical properties. Additionally, four studies extracted cluster cores or representative samples to interpret the characteristics of the clusters. For example, \citet{chen2018data} extracted cluster cores using KNN and \citet{Zhang_2015} constructed a transition matrix of Markov chains to identify and connect patient states that occur with high probability and frequency, within a cluster.

Evaluation of clustering results was commonly achieved by comparing cluster quality (e.g. inter- and intra-cluster distance, mutual information, or Rand index) against clusters generated via benchmark methods, and by verifying the meaningfulness of the clusters against external information or domain knowledge (such as patient demographics and clinical best practices). Another evaluation technique was to use cluster membership as a feature in the classification algorithm, using it to optimise the classification accuracy of data and to indicate that meaningful clusters were formed during the analysis. 

\begin{table*}[]
	\caption{Cluster analysis techniques}
	\label{tab:clustering-stagesAndTechniques}
	\begin{tabular}{|p{0.15\textwidth}|p{0.45\textwidth}|p{0.3\textwidth}|}
		\hline
	\textbf{Step} & \textbf{Technique} & \textbf{Studies}
	\\ \hline
		\multirow{9}{*}
		{Data Representation}
		& 
		Clinical pathway as a string &	\cite{	Zhang_2015,apunike2020analyses,bose2009trace,prokofyeva2019application,aspland2021modified}
		\\ \cline{2-3} 
		&
		Vector of diagnosis codes by occurrence frequency & \cite{doshi-velez2014comorbidity,roque2011using}
		\\ \cline{2-3} 	
		&
		Treatment record set sequence &	\cite{chen2018data,chen2020fusion}
		\\ \cline{2-3} 
		& 
		Vector of patient state &	\cite{johns2020clustering,hompes2015discovering}
				\\ \cline{2-3} 
		&
		Disease co-occurrence vector &	\cite{sideris2016flexible}
		\\ \cline{2-3} 
		& 
		Diagnosed disease trajectory & \cite{jensen2014temporal}
		\\ \cline{2-3} 	
		&

		Cancer metastasis dynamic network &	\cite{chen2009cancer}
		\\ \cline{2-3} 
		&
	  Patient flow dynamic network &	\cite{bean2017network}
	  		\\ \cline{2-3} 	
	  & 
	  Drug use duration statistic vector (novel) & \cite{chen2019mining}	
		\\ \hline	
		
		\multirow{13}{*}
		{Similarity Measure}
		&
		Euclidean distance & \cite{doshi-velez2014comorbidity,chen2019mining,bean2017network}
		\\ \cline{2-3} 
		&	
		Jaccard score & \cite{sideris2016flexible,jensen2014temporal}
	\\ \cline{2-3} 
&
		Cosine similarity
& \cite{roque2011using,hompes2015discovering}
\\ \cline{2-3} 
&
		Levenshtin distance
& \cite{apunike2020analyses}
	\\ \cline{2-3} 
&		
		Damerau–Levenshtein distance & \cite{prokofyeva2019application}
	\\ \cline{2-3} 
&
		Metastasis conditional incidence (hazard) functions
& \cite{chen2009cancer}
	\\ \cline{2-3} 
&

		Longest common subsequence
& \cite{Zhang_2015}
		\\ \cline{2-3} 
	&	
		Maximal Repeat Alphabet Feature Set (novel)
& \cite{bose2009trace}
			\\ \cline{2-3} 
		&

		Fusion of 3 novel similarity measures- content sequence and duration view (novel) & \cite{chen2020fusion}
	\\ \cline{2-3} 
&	
		 Modified Needleman–Wunsch algorithm  (novel)& \cite{aspland2021modified}
		 	\\ \cline{2-3} 
		 &
		 Markov theory based measure  (novel)& \cite{chen2018data}
		 	\\ \cline{2-3} 
		 &
		Ordinal edit distance (novel)& \cite{johns2020clustering}
		\\ \hline
		
		\multirow{2}{*}
		{Clustering Algorithm}
		&

		Hierarchical agglomerative clustering with Ward's method	 & \cite{doshi-velez2014comorbidity,sideris2016flexible,chen2009cancer,johns2020clustering,bose2009trace,prokofyeva2019application,zhang2018patient2vec}
		\\ \cline{2-3} 
		
				&
		Hierarchical clustering based on complete linkage	 & \cite{apunike2020analyses}
		\\ \cline{2-3} 
		
						&
		Hierarchical clustering based on average linkage	 & \cite{roque2011using}
		\\ \cline{2-3} 
		&
		Affinity propagation	 & \cite{chen2018data,chen2019mining}
		\\ \cline{2-3} 
		&
		Markov Cluster Algorithm  & \cite{hompes2015discovering,jensen2014temporal}
	 		\\ \cline{2-3}  &
		
		Spectral Clustering & \cite{chen2020fusion}
	 		\\ \cline{2-3} &
				
	K-means & \cite{chen2020fusion}
		\\ \cline{2-3} &
	
	K-medoids clustering & \cite{aspland2021modified}
		\\ \cline{2-3} &
	PCA based clustering & \cite{bean2017network}
	\\ \hline
	
\multirow{2}{*}
	{Interpretation}
	&
	Descriptive analytics and visualisations & \cite{doshi-velez2014comorbidity,roque2011using,chen2009cancer,apunike2020analyses,johns2020clustering,chen2020fusion,chen2019mining,bean2017network,hompes2015discovering} 
	\\ \cline{2-3} &
	
	Extract  cluster core/representative samples & \cite{chen2018data,apunike2020analyses,Zhang_2015,prokofyeva2019application} 
	\\ \cline{2-3} &

Compare with clinical guidelines & \cite{chen2020fusion}
	\\ \hline
	
	\multirow{2}{*}
	{Evaluation}
	&
  Cluster quality & \cite{johns2020clustering,bose2009trace,chen2020fusion,chen2019mining,aspland2021modified} 
	\\ \cline{2-3} &	
	Verifying interpretability of clusters & \cite{chen2018data,chen2019mining,Zhang_2015,jensen2014temporal,hompes2015discovering}
			\\ \cline{2-3} &
				Using cluster membership as a feature for classification & \cite{sideris2016flexible,chen2020fusion}
	\\ \hline
	
	\end{tabular}
\end{table*}
\xspace 

\subsubsection{Network Analysis}
\label{subsec:network_analysis}

\begin{figure}[h!]
	\centering
	\tikzstyle{decision} = [diamond, draw, fill=blue!20,
    text width=4.5em, text badly centered, node distance=2.5cm, inner sep=0pt]
	\tikzstyle{block} = [rectangle, draw, fill=blue!20,
		text width=7em, text centered, rounded corners, minimum height=4em]
	\tikzstyle{line} = [draw, very thick, color=black!50, -latex']
	\tikzstyle{cloud} = [draw, ellipse,fill=red!20, node distance=2.5cm,
		minimum height=2em]
	\resizebox{\columnwidth}{!}{
		\begin{tikzpicture}[scale=1, node distance = 3.5cm, auto]
			\node [block] (init) {Node Representation};
			\node [block, right of=init] (Edge) {Edge Representation};
			\node [block, right of=Edge] (Design) {Network Design};
			\node [block, below=0.5cm of Design] (Analysis) {Network Analysis};
			\node [block, left of=Analysis] (Interpretation) {Interpretation};
			\node [block, left of=Interpretation] (Usage) {Usage};
			\path [line] (init) -- (Edge);
			\path [line] (Edge) -- (Design);
			\path [line] (Design) -- (Analysis);
			\path [line] (Analysis) -- (Interpretation);
			\path [line] (Interpretation) -- (Usage);
		\end{tikzpicture}
	}

	\caption{Steps followed in network analysis-based studies}
	\label{fig:networkAnalysis}
\end{figure}
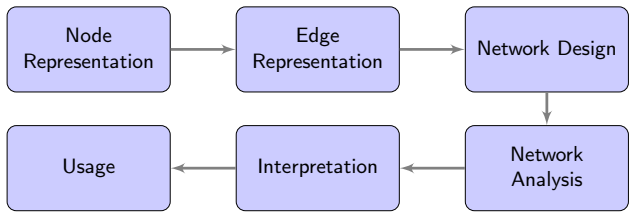

A total of fourteen\xspace studies use network analysis to represent and analyse health service patterns.
Network analysis based studies were observed to follow five steps as shown in Fig. \ref{fig:networkAnalysis}. 
The specific techniques and studies associated with each step are listed in Table \ref{tab:network-stagesAndTechniques}. 

In network analysis studies, nodes most often represent diseases (six studies) or clinical events (five studies), while edges most often represent co-occurrence of records (six studies) or a temporal relationship (five studies). Certain studies have reduced the network dimensions by means such as grouping the diseases in co-morbidity networks using an ontological hierarchy of ICD-9-CM codes. Co-occurrence measures used to define relationships in networks varied from simple measures such as co-occurrence frequency to TF-IDF and statistical equations such as pairwise association analysis using an chi-square test and correlation. We observed that all network designs are weighted, with five directed networks. Three studies each have utilised temporal and dynamic networks designs as well. Two studies~\cite{chen2009cancer,bean2017network} were observed to utilise unique network representations to look into cancer metastasis spread and patient flow in hospital wards.

Various techniques have been utilised to analyse and interpret the resulting network as listed in Table \ref{tab:network-stagesAndTechniques}, with a majority of studies using visualisations and statistical interpretations of network properties. Some studies have utilised network properties such as hubs, dense cores, communities and degree variability to understand and interpret network behaviour.

Network representations were used widely for discriprive analytics and to generate measures for downstream analysis tasks such as clustering or risk prediction. Descriptive analysis, pattern detection and observing emergent behaviour over dynamic networks were also objectives of some studies.

A noteworthy observation is that network analysis studies often do not include an evaluation step. We attribute this to the inherent explainability of network models and simple process of network representation and design, which makes the network analysis results intuitive, traceable and any interpretation and usage easily explainable.

\begin{table*}[]
	\caption{Network analysis techniques}
	\label{tab:network-stagesAndTechniques}
	\begin{tabular}{|p{0.16\textwidth}|p{0.44\textwidth}|p{0.3\textwidth}|}
		\hline
		\textbf{Step} & \textbf{Technique} & \textbf{Studies}
		\\ \hline

		\multirow{5}{*}
		{Node Representation}
		& 
		Disease (Co-morbidity network) &	\cite{roque2011using,sideris2016flexible,steinhaeuser2009network,hanauer2013modeling,glicksberg2016comparative,kannan2016conditional}
		\\ \cline{2-3}  &
		Patient & \cite{roque2011using,steinhaeuser2009network,ochoa2022graph}		
			\\ \cline{2-3} 	&
		Clinical event & \cite{liu2015temporal,Dong_2021,Kushima_2019,Zhang_2017,Wang_2021}
				\\ \cline{2-3} 	&
		Sites of metastasis & \cite{chen2009cancer}
		\\ \cline{2-3}  &	
		Hospital wards & \cite{bean2017network}
		\\ \hline
		
		\multirow{4}{*}
		{Edge Representation}
		& 
		Measure of co-occurrence in records &	\cite{roque2011using,sideris2016flexible,steinhaeuser2009network,hanauer2013modeling,Dong_2021,Wang_2021,ochoa2022graph}
		\\ \cline{2-3} 
		&
		Temporal relationship & \cite{liu2015temporal,bean2017network,kannan2016conditional,Kushima_2019,Zhang_2017}
		\\ \cline{2-3} 
		&
		Risk of developing subsequent disease & \cite{chen2009cancer,steinhaeuser2009network,glicksberg2016comparative}
			    \\ \cline{2-3} 
		&
	 	Causal information fraction & \cite{kannan2016conditional}
	\\ \hline
		
		\multirow{4}{*}
		{Network Design}
		& 
		Weighted & 	\cite{roque2011using,sideris2016flexible,chen2009cancer,bean2017network,steinhaeuser2009network,hanauer2013modeling,glicksberg2016comparative,kannan2016conditional,liu2015temporal,Dong_2021,ochoa2022graph}
		\\ \cline{2-3} 
		&
		Directed &	\cite{chen2009cancer,bean2017network,hanauer2013modeling,glicksberg2016comparative, kannan2016conditional,liu2015temporal,Dong_2021,Kushima_2019}
		\\ \cline{2-3} 
		&
		Temporal & \cite{chen2009cancer,bean2017network,hanauer2013modeling,Zhang_2017}
		\\ \cline{2-3} 
		&
		Dynamic &	\cite{chen2009cancer,bean2017network,steinhaeuser2009network}
		\\ \hline
		
		\multirow{10}{*}
		{Interpretation}
		& 
		Visualisation & \cite{roque2011using,chen2009cancer,hanauer2013modeling,glicksberg2016comparative,kannan2016conditional,liu2015temporal,Kushima_2019,Zhang_2017}
		\\ \cline{2-3} 
		&
		Statistical properties of network &	\cite{roque2011using,chen2009cancer,steinhaeuser2009network,glicksberg2016comparative,kannan2016conditional,Wang_2021}
		\\ \cline{2-3} 
		&
		Hubs & \cite{steinhaeuser2009network,glicksberg2016comparative}
					\\ \cline{2-3} 	&
		Dense core & \cite{roque2011using,steinhaeuser2009network}
			\\ \cline{2-3} 	&
		Community detection & \cite{steinhaeuser2009network,kannan2016conditional}	
			\\ \cline{2-3} 	&
		Long-range edges & \cite{steinhaeuser2009network}
		\\ \cline{2-3} 	&
		Degree stability over time & \cite{bean2017network}
			\\ \cline{2-3} 	&
		Edge weight variability over time	& \cite{bean2017network}
		\\ \cline{2-3} 	&
		Extract temporal phenotypes (representative network) & \cite{liu2015temporal}
		\\ \cline{2-3} 	&
		Random walks & \cite{Dong_2021}%
		\\ \hline

		\multirow{4}{*}
		{Usage}
		& 
		Descriptive analytics & \cite{roque2011using,chen2009cancer,bean2017network,steinhaeuser2009network,hanauer2013modeling,glicksberg2016comparative,liu2015temporal,Kushima_2019,Wang_2021}
		\\ \cline{2-3} 
		&
		Generate measures for downstream analysis &	\cite{roque2011using,sideris2016flexible,chen2009cancer,steinhaeuser2009network,kannan2016conditional,liu2015temporal,ochoa2022graph}
		\\ \cline{2-3} 
		&
		Pattern detection & \cite{glicksberg2016comparative,kannan2016conditional,liu2015temporal,Dong_2021,Kushima_2019}
		\\ \cline{2-3} 	&
		Observe emergent behaviour over time & \cite{steinhaeuser2009network}
		\\ \hline

		\end{tabular}
	\end{table*}
\xspace 

\subsubsection{Markov Models}\label{sec:markovmodels}

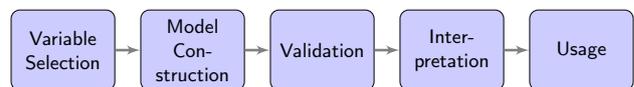
\begin{figure}[h!]
	\centering
	\tikzstyle{decision} = [diamond, draw, fill=blue!20,
    text width=4.5em, text badly centered, node distance=2.5cm, inner sep=0pt]
	\tikzstyle{block} = [rectangle, draw, fill=blue!20,
		text width=4.5em, text centered, rounded corners, minimum height=4em]
	\tikzstyle{line} = [draw, very thick, color=black!50, -latex']
	\tikzstyle{cloud} = [draw, ellipse,fill=red!20, node distance=2.5cm,
		minimum height=2em]
	\resizebox{\columnwidth}{!}{  
		\begin{tikzpicture}[scale=1, node distance = 2.1cm, auto]
			\node [block] (init) {Variable Selection};
			\node [block, right of=init] (Construction) {Model Construction};
			\node [block, right of=Construction] (Validation) {Validation};
			\node [block, right of=Validation] (Interpretation) {Inter-\\pretation};
			\node [block, right of=Interpretation] (Usage) {Usage};
			\path [line] (init) -- (Construction);
			\path [line] (Construction) -- (Validation);
			\path [line] (Validation) -- (Interpretation);
			\path [line] (Interpretation) -- (Usage);
		\end{tikzpicture}
	}

	\caption{Steps followed in Markov-based studies}
	\label{fig:markov}
\end{figure}

We identified seven\xspace studies that follow a Markov model-based analysis for health service pattern discovery in AHR. Markov model-based studies follow five steps as illustrated in Fig.~\ref{fig:markov}, and Table~\ref{tab:markov-stagesAndTechniques} lists the techniques applied in each step.

The most common way that state is represented in the Markov modelling-based studies is as either patient health state or a sequence of events and dynamic features. The Markov models applied in the studies are often constructed using more complex variations of the Markov model, such as a hidden Markov model or other novel extensions. This is to account for some hidden complexity that isn't captured in AHRs~\cite{leontjeva2016complex} or to incorporate some domain knowledge into the model~\cite{huang2018probabilistic}. 

Evaluation is based on domain knowledge and expert consultation, comparing performance with baseline methods and by numerical simulations. Probability-based descriptive analytics and visualisation are used to interpret Markov model results, but are used for different objectives such as understanding treatment pathways, for preprocessing and interpretation, and for generating features for downstream analytics.

It is important to note that healthcare is inherently a non-Markovian process, since the events recorded in AHR datasets contain long-term dependencies and do not explicitly depend on the previous state. For example, a routine admission with irrelevant medical information would destroy the illness memory, especially for chronic conditions \cite{pham2017predicting}. This should be considered carefully when applying Markov models for health service pattern discovery.

\begin{table*}[]
	\caption{Markov Modelling techniques}
	\label{tab:markov-stagesAndTechniques}
	\begin{tabular}{|p{0.2\textwidth}|p{0.5\textwidth}|p{0.2\textwidth}|}
		\hline
		\textbf{Step} & \textbf{Technique} & \textbf{Studies}
		\\ \hline
		

		\multirow{2}{*}
		{State Representation}
		& 
		Patient health state & 
		\cite{maass2020markov,baker2017process,Bueno_2018}
		\\ \cline{2-3}  &
		Sequences of events and dynamic features & 
		\cite{Zhang_2015,huang2018probabilistic,leontjeva2016complex}	
		\\ \hline	

		\multirow{5}{*}
		{Modelling Technique}
		& 
		Markov chain  & 
		\cite{Zhang_2015}
		\\ \cline{2-3}  &
		Finite horizon markov decision process & 
		\cite{maass2020markov}
		\\ \cline{2-3}  &
		Markov model  & 
		\cite{baker2017process}
		\\ \cline{2-3}  &
		Hidden Markov Model (HMM) &	
		\cite{leontjeva2016complex,Bueno_2018}
		\\ \cline{2-3}  &
		Novel extension of HMM & 
		\cite{huang2018probabilistic}	
		\\ \hline		

		\multirow{3}{*}
		{Evaluation}
		&
		Domain knowledge and expert consultation  &	
		\cite{Zhang_2015,baker2017process,Bueno_2018}
		\\ \cline{2-3}  &
		 Compare performance with baseline methods  & 
		 \cite{huang2018probabilistic,leontjeva2016complex}	
		\\ \cline{2-3}  &
		Numerical simulations &	
		\cite{maass2020markov}
		\\ \hline	

		\multirow{2}{*}
		{Interpretation}
		& 
		Probability based descriptive analysis & 
		\cite{Zhang_2015,maass2020markov,baker2017process,huang2018probabilistic}	
		\\ \cline{2-3}  &
		Visualisation &  
		\cite{Zhang_2015,maass2020markov,huang2018probabilistic}	
		\\ \hline

		\multirow{4}{*}
		{Usage}
		& 
		Understand and improve treatment pathways & 
		\cite{maass2020markov,baker2017process,huang2018probabilistic}	
		\\ \cline{2-3}  &
		Preprocessing and interpretation & 
		\cite{Zhang_2015}	
		\\ \cline{2-3}  & 
		Feature for downstream analytics & 
		\cite{leontjeva2016complex}	
		\\ \cline{2-3}  & 
		Correlated with medical-oriented outcomes & 
		\cite{Bueno_2018}	
		\\ \hline

	\end{tabular}
\end{table*}
\xspace 

\subsubsection{Regression-based Techniques}\label{sec:linearregression}

\begin{figure}[h!]
	\centering
	\tikzstyle{decision} = [diamond, draw, fill=blue!20,
    text width=4.5em, text badly centered, node distance=2.5cm, inner sep=0pt]
	\tikzstyle{block} = [rectangle, draw, fill=blue!20,
		text width=4.5em, text centered, rounded corners, minimum height=4em]
	\tikzstyle{line} = [draw, very thick, color=black!50, -latex']
	\tikzstyle{cloud} = [draw, ellipse,fill=red!20, node distance=2.5cm,
		minimum height=2em]
	\resizebox{\columnwidth}{!}{
		\begin{tikzpicture}[scale=1, node distance=2.1cm, auto]
			\node [block] (init) {Variable Selection};
			\node [block, right of=init] (Regression) {Regression analysis};
			\node [block, right of=Regression] (Evaluation) {Evaluation};
			\node [block, right of=Evaluation] (Interpretation) {Inter-\\pretation};
			\node [block, right of=Interpretation] (Usage) {Usage};
			\path [line] (init) -- (Regression);
			\path [line] (Regression) -- (Evaluation);
			\path [line] (Evaluation) -- (Interpretation);
			\path [line] (Interpretation) -- (Usage);
		\end{tikzpicture}
	}
	\caption{Steps followed in regression-based studies}
	\label{fig:regression}
\end{figure}
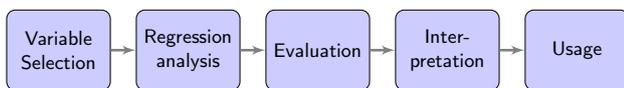

We identified six\xspace studies that use regression-based techniques to analyse health service patterns, following the five steps illustrated in Fig. \ref{fig:regression}. Table \ref{tab:regression-stagesAndTechniques} lists the techniques used in different studies for the four steps followed by variable selection.

All of the regression-based studies use similar variable characteristics to select which variables would be used in the analysis, such as whether they are dependent/independent, and which indexes or metrics are used to measure them. 

The most common regression model used was multivariate logistic regression, but other variations were also observed. Similarly, all of the regression-based studies evaluate their findings using statistical significance and sensitivity analysis, as well as using other evaluation techniques and validating results with domain knowledge where applicable~\cite{glicksberg2016comparative}. \citet{te2019alignment} used an additional validation step to ensure survival difference is independent of the age-at-diagnosis by using the multivariable Cox proportional hazard method.

All studies use descriptive statistical analytics and/or visualisations to interpret their results. Finally, each of the studies use the regression analysis to better understand clinical practice and associations between AHR variables. \citet{glicksberg2016comparative} quantifies the risk associated with diseases by using their regression analysis as a preprocessing step for edge weights in a disease network.

\begin{table*}[]
	\caption{Regression techniques}
	\label{tab:regression-stagesAndTechniques}
	\begin{tabular}{|p{0.2\textwidth}|p{0.5\textwidth}|p{0.2\textwidth}|}
		\hline 
		\textbf{Step} & \textbf{Technique} & \textbf{Studies}
		\\ \hline
		
		\multirow{3}{*}
		{Analysis Technique}
		& 
		Multivariate logistic regression & \cite{glicksberg2016comparative,roder2021female,shahabi2020differences,sun2022applying}	
		\\ \cline{2-3}  &
		Ordered logistic regression &	\cite{te2019alignment}
		\\ \cline{2-3}  &

		Proportional hazards regression & \cite{roder2021female}	
		\\ \hline
		
		\multirow{4}{*}
		{Evaluation}
		& 
		Statistical significance and sensitivity analysis & \cite{glicksberg2016comparative,roder2021female,te2019alignment,shahabi2020differences,sun2022applying} 
				\\ \cline{2-3}  &
		Confidence level & \cite{roder2021female,te2019alignment,shahabi2020differences}
		\\ \cline{2-3}  &		
		Multivariate Cox proportional hazard method &	\cite{te2019alignment} 
		\\ \cline{2-3}  &
		Against known data and clinical expectations & \cite{glicksberg2016comparative}	 
		\\ \hline
	
		\multirow{2}{*}	
		{Interpretation}
		& 
		Descriptive statistical analytics with tables &	\cite{glicksberg2016comparative,roder2021female,te2019alignment,shahabi2020differences,sun2022applying} 
		\\ \cline{2-3}  &
		Visualisation & \cite{glicksberg2016comparative,te2019alignment,sun2022applying} 
		\\ \hline

		\multirow{3}{*}
		{Usage}
		& 
		Discuss clinical implications & \cite{glicksberg2016comparative,roder2021female,shahabi2020differences,sun2022applying}
		\\ \cline{2-3}  &
		Identify new associations &	\cite{roder2021female,te2019alignment,shahabi2020differences}
		\\ \cline{2-3}  &
		Preprocessing step & \cite{glicksberg2016comparative}	
		\\ \hline		
		\end{tabular}
\end{table*}
\xspace 

\subsubsection{Other Techniques}\label{sec:othertechniques}

This section summarises the studies that utilise less common analytics methods. We briefly describe how these techniques are applied according to the features used, modelling techniques, and, where it was observed, methods for evaluation, interpretation, and usage. Our synthesis of the approaches to using these techniques may be limited in some cases due to the modest number of studies that utilise these methods. 

\begin{table*}[h]
	\caption{Sequence Mining}
	\label{tab:other-sequencemining}
	\begin{tabular}{|p{0.2\textwidth}|p{0.6\textwidth}|p{0.1\textwidth}|}
		\hline
		\textbf{Step} & \textbf{Technique} & \textbf{Studies}
		\\ \hline
		\multirow{1}{*}
		{Features Used}
		&
		Treatment data & \cite{kaur2020time,Estiri_2020,Vincent_Paulraj_2021}
		\\ \hline					
		\multirow{3}{*}
		{Modelling Technique}
		&
		Time range based sequence mining &	
		\cite{kaur2020time}
		\\ \cline{2-3}  &
		Transitive Sequential Pattern Mining (novel) &	
		\cite{Estiri_2020}
		\\ \cline{2-3}  &
		Sequential Relational n-Disease Pattern (SERENDIP) (novel) &	
		\cite{Vincent_Paulraj_2021}
		\\ \hline		
		\multirow{2}{*}
		{Evaluation}
		&
		Compare prediction accuracy with baseline methods & 
		\cite{kaur2020time,Estiri_2020}
		\\ \cline{2-3}  &
		Prediction accuracy & 
		\cite{Vincent_Paulraj_2021}
		\\ \hline		
		\multirow{1}{*}
		{Interpretation}
		&
		Identify significance of various attributes & 
		\cite{Vincent_Paulraj_2021,kaur2020time,Estiri_2020}
		\\ \hline
		\multirow{1}{*}
		{Usage}
		&
		Features for downstream analytics & 
		\cite{kaur2020time,Estiri_2020}	
		\\ \hline		

	\end{tabular}
\end{table*}

\begin{table*}[h]
	\caption{Topic Modelling}
	\label{tab:other-topicmodelling}
	\begin{tabular}{|p{0.2\textwidth}|p{0.5\textwidth}|p{0.2\textwidth}|}
		\hline

		\textbf{Step} & \textbf{Technique} & \textbf{Studies}
		\\ \hline


		\multirow{1}{*}
		{Features Used}
		& 
		Patient features and treatment behaviors & \cite{huang2015mining,Huang_2015a}	
		\\ \cline{2-3}  &
		Pathway as a text string &	\cite{prokofyeva2019application}		
		\\ \hline		
		\multirow{1}{*}
		{Modelling Technique}
		& 
		Latent Dirichlet Allocation (LDA)  &	\cite{huang2015mining}
		\\ \cline{2-3}  &
		Additive Regularisation of Topic Models (ARTM) &	\cite{prokofyeva2019application}	
		\\ \cline{2-3}  &
		Treatment Pattern Model (TPM) (novel) &
		\cite{Huang_2015a}
		\\ \hline		

		\multirow{1}{*}
		{Evaluation}
		& 
		Compare performance with baseline methods& \cite{huang2015mining}	
		\\ \cline{2-3}  &
		Perplexity score and sparsity for optimal topic numbers  &	\cite{prokofyeva2019application}		
		\\ \hline	
		\multirow{1}{*}
		{Interpretation}
		& 
		Visualisation &  \cite{huang2015mining}	
		\\ \cline{2-3}  &
		Probability based descriptive analysis & \cite{prokofyeva2019application}
		\\ \cline{2-3}  &
		
		Discover underlying treatment patterns & 
		\cite{Huang_2015a}
		\\ \hline

		\multirow{1}{*}
		{Usage}
		&
		Understand clinical pathways & \cite{prokofyeva2019application,huang2015mining,Huang_2015a}	
		\\ \hline
	\end{tabular}
\end{table*}

\begin{table*}[h]
	\caption{Ensemble Learning}
	\label{tab:other-ensemblelearning}
	\begin{tabular}{|p{0.2\textwidth}|p{0.5\textwidth}|p{0.2\textwidth}|}
		\hline

		\textbf{Step} & \textbf{Technique} & \textbf{Studies}
		\\ \hline


		\multirow{3}{*}
		{Features Used}
		&  
		Clinical events & 
		\cite{Boland_2015,Maali_2018} 
		\\ \cline{2-3}  &
		Administrative events &
		\cite{Wolff_2020,Boland_2015,Maali_2018}
		\\ \cline{2-3}  &
		Patient demographics &
		\cite{Wolff_2020,Maali_2018}
		\\ \hline	
		
		\multirow{3}{*}
		{Modelling Technique}
		& 
		Random Forest &	
		\cite{Boland_2015}
		\\ \cline{2-3}  &
		Stochastic Gradient Boosting (GBM) & 
		\cite{Wolff_2020,Maali_2018}
		\\ \cline{2-3}  &
		Multiple logistic regression & 
		\cite{Wolff_2020,Maali_2018}
		\\ \hline
		\multirow{2}{*}
		{Evaluation}
		& 
		Prediction accuracy & 
		\cite{Wolff_2020,Boland_2015,Maali_2018}
		\\ \cline{2-3}  &
		Compare performance with baseline methods & 
		\cite{Wolff_2020}
		\\ \hline
		\multirow{3}{*}
		{Interpretation}
		& 
		Measure phenotype severity & 	
		\cite{Boland_2015}
		\\ \cline{2-3}  &
		Predict psychiatric hospital care &
		\cite{Wolff_2020}
		\\ \cline{2-3}  &
		Predict hospital readmission &
		\cite{Maali_2018}
		\\ \hline
		\multirow{2}{*}
		{Usage}
		& 
		Outcome assessment  & 
		\cite{Boland_2015}
		\\ \cline{2-3}  &
		Decision support & 
		\cite{Wolff_2020,Maali_2018}
		\\ \hline		
	\end{tabular}
\end{table*}

\begin{table*}[h]
	\caption{Bayesian Networks}
	\label{tab:other-bayesiannetworks}
	\begin{tabular}{|p{0.2\textwidth}|p{0.6\textwidth}|p{0.1\textwidth}|}
		\hline
		\textbf{Step} & \textbf{Technique} & \textbf{Studies}
		\\ \hline


		\multirow{3}{*}
		{Features Used}
		& 
		{Patient health state} & 
		\cite{Nagrecha_2017}
		\\ \cline{2-3}  &
		Demographic, diagnosed-based and prior-utilization variable & 
		\cite{wang2020survivability}	
		\\ \cline{2-3}  &
		{EHR events corresponding to patient risk factors} & 
		\cite{Weiss_2013}
		\\ \hline	
		
		\multirow{2}{*}
		{Modelling Technique}
		& 
		Bayesian Networks &	
		\cite{Nagrecha_2017,wang2020survivability}
		\\ \cline{2-3}  &
		Multiplicative-Forest Point Processes (MFPPs) (novel) & 
		\cite{Weiss_2013}
		\\ \hline

		\multirow{2}{*}
		{Evaluation}
		& 
		Prediction accuracy & 
		\cite{Nagrecha_2017,wang2020survivability,Weiss_2013}	
		\\ \cline{2-3}  &
		Compare performance with baseline methods & 
		\cite{wang2020survivability}	
		\\ \hline		

		\multirow{3}{*}
		{Interpretation}
		& 
		Probability density function based & 
		\cite{wang2020survivability}	
		\\ \cline{2-3}  &
		{Likelihood of disease} & 
		\cite{Nagrecha_2017,Weiss_2013}
		\\ \cline{2-3}  &
		{Disease progression trajectory} & 
		\cite{Nagrecha_2017}
		\\ \hline

		\multirow{3}{*}
		{Usage}
		& 
		Identify disease risk factors & \cite{wang2020survivability}	
		\\ \cline{2-3}  &
		Predict survivability & \cite{wang2020survivability,Weiss_2013}	
		\\ \cline{2-3}  &
		{Improve understanding of diagnosis severity} & 
		\cite{Nagrecha_2017}
		\\ \hline
	\end{tabular}
\end{table*}

\begin{table*}[h]
	\caption{Association Rule Mining}
	\label{tab:other-associationrulemining}
	\begin{tabular}{|p{0.2\textwidth}|p{0.6\textwidth}|p{0.1\textwidth}|}
		\hline
		\textbf{Step} & \textbf{Technique} & \textbf{Studies}
		\\ \hline
		\multirow{2}{*}
		{Features Used}
		& 
		Clinical Events &  \cite{Vincent_Paulraj_2021}
		\\ \cline{2-3}  &
		{Diagnosis codes} & 
		\cite{Nguyen_2015}
		\\ \hline	

		\multirow{2}{*}
		{Modelling Technique}
		& 
		Sequential, Relational, n-Disease Pattern (novel) &	 \cite{Vincent_Paulraj_2021}
		\\ \cline{2-3}  &
		{A priori algorithm} & 
		\cite{Nguyen_2015}
		\\ \hline

		\multirow{2}{*}
		{Interpretation}
		& 
		Frequent item sets &  \cite{Vincent_Paulraj_2021}
		\\ \cline{2-3}  &
		{Pattern discovery} & 
		\cite{Nguyen_2015}
		\\ \hline
		\multirow{2}{*}
		{Usage}
		& 
		Multi-morbidity disease prediction & \cite{Vincent_Paulraj_2021}
		\\ \cline{2-3}  &
		{Discover toxicity patterns} &
		\cite{Nguyen_2015}
		\\ \hline		
	\end{tabular}
\end{table*}

\begin{table*}[h]
	\caption{Temporal Signature Mining}
	\label{tab:other-temporalsignaturemining}
	\begin{tabular}{|p{0.2\textwidth}|p{0.6\textwidth}|p{0.1\textwidth}|}
		\hline
		\textbf{Step} & \textbf{Technique} & \textbf{Studies}
		\\ \hline
		\multirow{2}{*}
		{Features Used}
		& 
		Patient encounters &	
		\cite{wang2012framework}		
		\\ \cline{2-3}  &
		{Diagnosis codes} & 
		\cite{Nguyen_2015}
		\\ \hline		

		\multirow{3}{*}
		{Modelling Technique}
		& 
		Novel Matrix Representation &	
		\cite{wang2012framework}
		\\ \cline{2-3}  &
		Extension of convolutional non-negative matrix factorisation &	
		\cite{wang2012framework}		
		\\ \cline{2-3}  &
		{A priori algorithm} & 
		\cite{Nguyen_2015}
		\\ \hline		

		\multirow{2}{*}
		{Evaluation}
		& 
		Clinical interpretations of identified signatures& 
		\cite{wang2012framework}	
		\\ \cline{2-3}  &
		Performance in detecting known patterns in synthetic data &	
		\cite{wang2012framework}		
		\\ \hline		
		\multirow{3}{*}
		{Interpretation}
		& 
		Visualisation & 
		\cite{wang2012framework}	
		\\ \cline{2-3}  &
		Clinical significance & 
		\cite{wang2012framework}	
		\\ \cline{2-3}  &
		{Pattern discovery} & 
		\cite{Nguyen_2015}
		\\ \hline	

		\multirow{2}{*}
		{Usage}
		&
		Understand latent event patterns of clinical significance & 
		\cite{wang2012framework}	
		\\ \cline{2-3}  &
		{Discover toxicity patterns} &
		\cite{Nguyen_2015}
		\\ \hline		
	\end{tabular}
\end{table*}

\begin{table*}[h]
	\caption{Dimensionality Reduction}
	\label{tab:other-dimensionalityreduction}
	\begin{tabular}{|p{0.2\textwidth}|p{0.6\textwidth}|p{0.1\textwidth}|}
		\hline
		\textbf{Step} & \textbf{Technique} & \textbf{Studies}
		\\ \hline
		\multirow{1}{*}
		{Features Used}
		& 
		Multiple similarity networks from treatment records&	
		\cite{chen2020fusion}		
		\\ \hline			
		\multirow{1}{*}
		{Modelling Technique}
		& 
		Laplacian Eigenmaps  &	\cite{chen2020fusion}
		\\ \hline			
		\multirow{1}{*}
		{Usage}
		&
		For downstream analysis& \cite{chen2020fusion}	
		\\ \hline		
	\end{tabular}
\end{table*}

\begin{description}
	\item[Sequence Mining] We identified five\xspace studies used sequence-based mining approaches. In these studies, authors cited the importance of capturing relationships between the order of events within AHRs. For example, \citet{kaur2020time} proposed a time range based sequence mining approach on treatment data to generate informative features for downstream analysis. The methods of these studies are further detailed in Table~\ref{tab:other-sequencemining}.

    \item[Topic Modelling] We identified four studies that use natural language processing-inspired topic modelling techniques (Latent Dirichlet Allocation (LDA) and Additive Regularisation of Topic Models (ARTM)), and one novel technique, Treatment Pattern Model (TPM). These studies identify health service patterns and use them for understanding clinical pathways. A common theme of these studies is the modelling of latent variables that are indirectly associated to the recorded variables in AHRs. The methods of these studies are further detailed in Table~\ref{tab:other-topicmodelling}.

	\item[Ensemble Learning] We identified three studies that use some form of ensemble-based learning techniques. These studies each pursue prediction-based tasks, often using multiple AHR attributes. The methods of these studies are further detailed in Table~\ref{tab:other-ensemblelearning}.

	\item[Bayesian Networks] Bayesian Networks were used in three studies to identify risk factors, and to predict survivability and progression of diseases. The probability-based interpretation of the model was cited as a particular advantage of Bayesian Networks. The methods of these studies are further detailed in Table~\ref{tab:other-bayesiannetworks}.
	
	\item[Association Rule Mining] Association rule mining was used on only two identified studies, typically for high-level AHR analysis -- such as pattern discovery and identifying frequent item sets -- and are further detailed in Table~\ref{tab:other-associationrulemining}.
	
	\item[Temporal Signature Mining] We identified two\xspace studies proposed temporal signature mining, which is distinct from sequence mining, since the time between events is considered in addition to the sequence-based nature of the data. \citet{wang2012framework} represent and analyse patient encounters through a novel matrix representation, and their approach is further detailed in Table~\ref{tab:other-temporalsignaturemining}.
	
	\item[Dimensionality Reduction] Only one\xspace study~\cite{chen2020fusion} used Laplacian Eigenmaps (LEs) to embed the adjacency graph of a large network into a low-dimensional space to conduct features for downstream analysis. See Table~\ref{tab:other-dimensionalityreduction}. Note that other approaches -- such as neural networks for representation learning (Section~\ref{sec:neuralnetworks}) or vector-based representations used for clustering (Section~\ref{sec:clustering}) -- may also effectively reduce the dimensionality of data. However, these are not included here since dimensionality reduction is not the primary motivation for such methods, nor was it the stated objective of the authors in such studies.   

\end{description}
\xspace 

%
%
%

\subsection{RQ2:  What are the applications of health service patterns in the identified studies?}\label{sec:results:RQ2}

We mapped the identified literature into eight key application areas that use machine learning techniques on EHR data. These applications are adapted from other applications recognised by existing EHR-based mining analytics surveys \cite{yadav2018mining,chen2017textual}, however we focus the scope of these applications specifically to AHR-based HSPM. 


Table~\ref{tab:ML-usage} shows which applications are pursued in each of the reviewed studies. We found that the most common applications were analysis of healthcare patterns and of medical trajectories. We believe this is for two reasons: 1. These applications are valuable to health informatics research, and 2. The characteristics of AHRs (e.g. their temporal nature) make these applications practical. 

Other common applications included: comorbidity analysis, healthcare guidelines, cohort identification, and risk prediction. These studies often use very distinct or unambiguous features of AHRs, such as diagnosis codes to predict risk~\cite{Zhang_2017,wang2020survivability} or identify comorbidities~\cite{doshi-velez2014comorbidity,sideris2016flexible}. Finally, only three studies pursue outlier detection, while only one study pursues intervention analysis. This may be due to a lack of distinct or unambiguous features in AHRs that correlate to high-level healthcare concepts such as interventions and anomalies. 

The remainder of this section describes each of these applications in some more detail.

\subsubsection*{Healthcare patterns} 
The most common application area was to understand the patterns and pathways in medical care that patients receive. This may include patterns of drug or procedure codes that are commonly prescribed (such as learned representations of medical-codes \cite{choi2016multi}), or care pathways across a healthcare system \cite{Zhang_2015}. In total, 27 studies attempted some form of healthcare pattern analysis. 

\subsubsection*{Medical trajectory}
Another common application, observed in 23 studies, was prediction of future medical events in the treatment sequence of a patient and prediction of other future phenomena such as outcome or readmission. For example, \citet{jensen2014temporal} use large, population-wide data registry to extract temporal disease trajectories.

\subsubsection*{Comorbidity analysis} 
This application is concerned with identifying patients that express two or more diseases or medical conditions at the same time. In one study, \citet{sideris2016flexible} develop a feature-extraction framework for identifying comorbidities in AHRs. We identified ten studies in total that applied AHR-based analytics for comorbidity analysis. 

\subsubsection*{Healthcare guidelines}
Nine studies sought to use AHRs to inform the development of evidence-based guidelines in healthcare. For example, \citet{roder2021female} observes correlations between country-of-birth and female breast cancer in New South Wales, Australia to draw implications for health-service delivery. 

\subsubsection*{Cohort identification}
Eight studies identify similar groups of patients within an AHR database according to an attribute or set of attributes. For example, \citet{roque2011using} identify patient cohorts within a population using a hierarchical stratification.

\subsubsection*{Risk prediction}
Seven studies quantify patient severity or risk based on attributes contained within AHRs. For example, \citet{wang2020survivability} determines a survivability prognosis for lung cancer patients by analysing relevant risk factors. 

\subsubsection*{Outlier detection}
This application relates to identifying patients who receive care that is not in line with that of similar patients. For example, \citet{hompes2015discovering} identifies variations in healthcare processes in event data from AHRs. 

\subsubsection*{Intervention analysis}
The final observed application area relates to evaluating the efficacy of healthcare interventions. In the one study that pursued this application, \citet{te2019alignment} used existing care pathways to observe survival outcomes among colon cancer patients.

\begin{table*}[width=.95\textwidth,cols=2]
	\caption{Applications of Health Service Patterns}\label{tab:ML-usage}
	\begin{tabular}{ll}
\toprule
          Application &                                                                                                                                                                                                                                                                                                                                                                                                                                                   Studies \\
\midrule
  Healthcare patterns & \cite{Zhang_2015,chen2018data,johns2020clustering,bose2009trace,chen2020fusion,prokofyeva2019application,Chambard_2021,steinhaeuser2009network,liu2015temporal,Dong_2021,Kushima_2019,choi2016multi,choi2017gram,guo2020comparative,zhang2018patient2vec,xu2018learning,Hong_2017,Lu_2021,steinberg2021language,Caruana_2022,shahabi2020differences,baker2017process,huang2018probabilistic,huang2015mining,wang2012framework,Estiri_2020,ochoa2022graph} \\
   Medical trajectory &                                                                                                                  \cite{chen2009cancer,jensen2014temporal,Mohan_Kumar_2020,steinhaeuser2009network,liu2015temporal,Zhang_2017,choi2018mime,pham2017predicting,li2020ccae,beaulieu2018mapping,Li_2020,Lu_2021,Wolff_2020,Li_2016,maass2020markov,leontjeva2016complex,Nagrecha_2017,Huang_2015a,Maali_2018,Weiss_2013,kaur2020time,Zheng_2021,Gerrard_2022} \\
 Comorbidity analysis &                                                                                                                                                                                                                                        \cite{sideris2016flexible,doshi-velez2014comorbidity,roque2011using,steinhaeuser2009network,glicksberg2016comparative,kannan2016conditional,Vincent_Paulraj_2021,Mohan_Kumar_2020,Bueno_2018,Vincent_Paulraj_2021} \\
Healthcare guidelines &                                                                                                                                                                                                                                                                                                           \cite{chen2018data,chen2019mining,aspland2021modified,bean2017network,jin2018treatment,roder2021female,du2019variance,Chambard_2021,Huang_2015} \\
Cohort identification &                                                                                                                                                                                                                                                                                                                         \cite{roque2011using,sideris2016flexible,apunike2020analyses,hanauer2013modeling,choi2016multi,Nguyen_2015,Boland_2015,Wang_2021} \\
      Risk prediction &                                                                                                                                                                                                                                                                                                                                                   \cite{Zhang_2017,Wang_2021,wang2020survivability,Nagrecha_2017,Boland_2015,Nguyen_2015,sun2022applying} \\
    Outlier detection &                                                                                                                                                                                                                                                                                                                                                                                                                      \cite{hompes2015discovering,Li_2016} \\
Intervention analysis &                                                                                                                                                                                                                                                                                                                                                                                                                                    \cite{te2019alignment} \\
\bottomrule
\end{tabular}
\xspace  
\end{table*}

\subsection{RQ3: What are the analytics activities associated with health service pattern mining?}\label{sec:results:RQ3}

\begin{figure*}[h!] 
	\centering
	\includegraphics[width=0.98\textwidth]{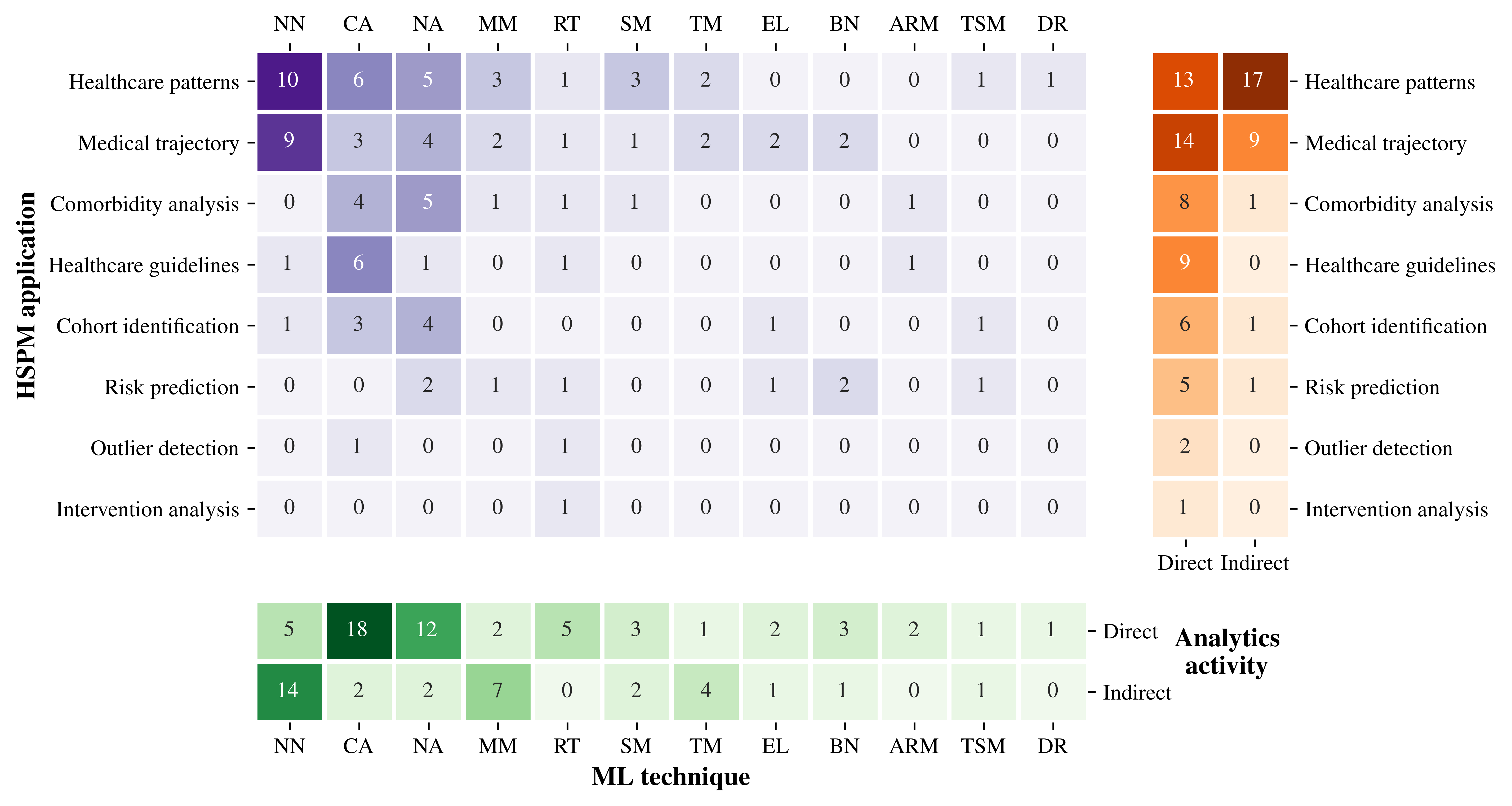}
	\caption{A 2D heatmap (purple) depicting how many studies (cell-value) utilise a given machine learning technique for each HSPM application. Also pictured are which studies undertook direct/indirect analytics activities, by both techniques (green) and applications (orange).} 
	\label{fig:rq3}
\end{figure*}

In sections \ref{sec:results:RQ1} and \ref{sec:results:RQ2}, we highlight the twelve machine learning techniques and eight HSPM applications identified in literature. This section looks at how these techniques are used in pursuit of these applications. First, we look at how often each technique is used for each application. Then, we look at the analytics activity undertaken in each study to gain further insight into the research. 

The 2D heatmap in Figure~\ref{fig:rq3} depicts the distribution of ML techniques and HSPM applications. It shows three major technique-related trends.

\begin{itemize}
	\item Clustering and network analysis are applied often and in many and diverse application areas, including healthcare patterns, medical trajectories, comorbidity analysis, healthcare guidelines, and cohort identification. A similar relationship was observed for network analysis, except it was used less often for healthcare guidelines and more often for risk prediction.
	\item Neural networks were applied almost exclusively to healthcare patterns and medical trajectory analysis. Also, Markov models, topic modelling, ensemble learning, Bayesian networks, sequence mining, and dimensionality reduction were also used almost exclusively for these same applications, though they were not as commonly used as neural networks. 
	\item There was no clear application-related trends for regression analysis, association rule mining, or temporal signature mining.
\end{itemize}

Furthermore, we also observed that the majority of studies (64\%) often use the associated machine learning techniques to directly pursue their desired application. A common example of a direct analytics activity is analysis of clusters or networks of drug or diagnosis codes. However, a significant number of studies (36\%) model healthcare attributes that are not recorded in AHRs indirectly using what is available in the data. For example, a common analytics activity is to learn more useful representations of medical codes that reveal latent medical- or healthcare-knowledge \cite{choi2016multi,choi2017gram,choi2018mime}. To distinguish between these two types of analytics activities, we refer to them as direct-analytics and indirect-analytics respectively, and define them as follows:

\begin{description}
	\item[Direct-analytics] Modelling or prediction of attributes that have clear definitions and observations.
	\item[Indirect-analytics] Modelling or prediction of attributes that are either conceptually abstract or are not directly observed in the data.
\end{description}

We distinguished each study identified in this review using these definitions, and present the results in Figure~\ref{fig:rq3} adjacent to the primary heatmap depicted. When distinguishing the techniques in this way, we find that the following techniques almost exclusively pursue direct-analytics tasks: clustering, network analysis, regression analysis, and association rule mining. In contrast, neural networks, Markov modelling, and topic modelling are used almost exclusively for indirect-analytics tasks. Furthermore, when distinguishing applications in this way, we find that indirect-analytics tasks are common for analysing healthcare patterns and medical trajectories, while direct-analytics tasks were common among every application. 

These findings suggest the choice of machine learning technique used should be informed by the analytics task. Furthermore, indirect-analytics is more applicable to the modelling of the latent and often abstract attributes not recorded in AHRs, and is common in modelling healthcare patterns and medical trajectories.

\subsection{RQ4: What are the limitations of existing machine learning techniques for health service pattern discovery?}\label{sec:results:RQ4}

This section discusses several important limitations regarding AHR and machine learning techniques for health service pattern discovery. 

\subsubsection{Limitations Associated with Data}\label{limit:data}
This subsection outlines seven relevant limitations that pertain to AHRs.

\subsubsection*{Data Quality} 

A common limitation we identify in literature is that of data quality, which can affect the performance, validity, and applicability of particular modelling approaches. Typical data quality issues include: missing values, incorrect values, inconsistent formatting, inconsistent reporting standards, or noise introduced into data from human or instrument errors.

Some machine learning techniques may be robust to particular data quality issues, however their applicability in such cases depends on the type and severity of the issues. For example, clustering was often applied to noisy datasets, but not for datasets with missing values. This is because many similarity measures used in clustering algorithms are tolerant of small variations in the data, but cannot compute similarity where data is missing \cite{johns2020clustering}. Other means of addressing data quality issues specific to AHRs include using ensemble-learning approaches and imputation algorithms for handling missing values~\cite{liu2020elmv,xue2019mixture} and class imbalance \cite{huda2016hybrid}.


\subsubsection*{Data Accessibility} 

AHRs contain information pertaining to a patient's health state and health service access, as well as personally identifiable attributes such as age, sex, or geographical locality. Access to AHRs is often restricted to limit the risk of the sensitive information being leaked or breached \cite{Ray_2006}. In a research context, these restrictions have implications for what parts of AHR-based research can be shared,  what findings can be revealed, and limit the ease with which research findings can be reproduced \cite{Myers2016}. 

\subsubsection*{Indirect Nature of Data} 

AHRs do not specifically seek to make an accurate recording of a patient's health state. The information recorded within AHRs is still valuable for health informatics since it indirectly relates to a patient's health state. For example, the primary purpose of billing codes recorded by hospitals or insurance providers is accounting, but these codes may be used to indirectly infer a patient's health state since they indicate which services the patient received. 

The key limitation of indirect data is noise and uncertainty, and it is critical to understand and quantify uncertainty when using indirect data. Using uncertain quantities as an evaluation criterion may lead to misleading or inaccurate research conclusions. For example, many of the studies reviewed in this paper explore the relationship between patient readmission (or lack thereof) and existing treatment patterns. In these studies, patient readmission is used as a proxy for the success or failure of a treatment. However, this is an indirect attribute for treatment success, since there are many reasons why patients may/may not be readmitted. 

Researchers should practise caution when using indirect relationships within AHRs to make causal assertions or predictions and, where possible, compute confidence intervals and estimate likelihood of any conclusions drawn from indirect means.

\subsubsection*{Fragmentation}  

The most accessible types of AHRs are those sourced from a single institution, such as hospital administration data, insurance claims data. Analysis of single-source data-sets exposes the indirect- and incomplete-data quality issues discussed above. To move beyond single-source AHRs, patients need to be linked across the various disparate single-source AHRs. We refer to this issue as AHR fragmentation. While many methods exist for linking entities across disparate databases, the aforementioned issue of data accessibility makes this difficult. The process of linking data further accentuates the data accessibility limitation, but linked databases may mitigate some data quality or completeness limitations. 

Single-source AHRs are often not sufficient for popu-lation-level health studies, since the relevant patient records may be distributed amongst multiple healthcare providers or across jurisdictional borders. The data fragmentation limitation significantly inhibits health informatics research at a population-level.

\subsubsection*{Irregular Temporal Data}

Recordings in AHRs are typically not done in a structured manner. For example, two patients with same diagnosis and treatment patterns may have distinct AHRs, since they may not follow the same visit frequency, or their institution or jurisdiction may observe different administrative recording practices. This is a significant limitation in AHR-based research, with some studies choosing to represent the temporal data simply as a sequence. Moreover, other studies -- especially those that calculate patient similarity using Clustering -- simply disregard temporal information entirely, aggregating information within discrete bins such as hospital visits instead. This indicates that the temporal information stored within AHRs is difficult to utilise, however not doing so risks losing valuable contextual insight from the data. 

Capitalising on the temporal information requires researchers to come up with similarity measures or feature representations that rely on assumptions that are context specific, hard to validate and generalise. We identify there is a need to benchmark such temporal representations of AHR data and report their performance across different machine learning tasks.

\subsubsection*{Data Quantity}

The quantity of data available in AHRs is another common limitation with regards to both the number of records (rows) and the number of attributes per record (columns). This problem is often a symptom of other limitations (such as fragmentation or data quality), but not always. For example, even population-level, comprehensively linked AHRs may not contain many records for patients with uncommon diseases. 
Insufficient data leads analytical methods to discover weak or inaccurate relationships or characteristics. Care should be taken to report on the confidence of predictions when AHRs are affected by this limitation to avoid weak confidences of results going unnoticed. 

It may be possible in some cases to rectify this limitation to a certain degree with data augmentation, or with synthetic
data. For example, \citet{steinberg2021language} develop a model to learn representations of medical records such that it does not suffer significantly from limited data. 

\subsubsection*{Incorporating domain knowledge} 

Domain-specific knowledge is often critical to making AHRs and AHR-based analytics interpretable, as
well as for organising data, pre-processing data, and for feature engineering. 

For example, we observe many studies use International Statistical Classification of Diseases and
Related Health Problems (ICD) \cite{ICD_10} codes. These codes are high-dimensional and not
human-readable, though they have human-readable definitions. Where feasible, these studies can
benefit from ontologies -- such as the Clinical Classification Software (CCS) \cite{CCS} -- to
reduce the sparsity of such codes sets by incorporating semantic, hierarchical structure. 

Aside from semantic structure, other code-grouping structures might also be clinically relevant. For
example, specific treatments may involve grouping of semantically distinct codes, and combinations
of specific procedure and diagnosis codes may indicate different degrees or severities of a
particular disease (e.g., early-stage vs terminal cancer). 

Another example of domain-specific knowledge is the use of clinical guidelines to inform the
construction of features or the design of models. For example, \citet{choi2016multi} use clinical
guidelines to inform the design of their model (e.g., grouping events by hospital admission), and
\citet{choi2017gram} use clinical guidelines to inform the construction of features (e.g.,
hierarchical features via CCS).

Finally, we note that domain-specific knowledge is often difficult to incorporate into AHR-based
analytics because the data is administrative and not in the healthcare domain. The semantic gap
between the administrative and clinical domains is significant, and designing methods to bridge this
gap (such as the CCS) requires significant domain expertise. Efforts to develop such domain-bridging
methods can vastly increase the clinical utility of AHR-based machine learning analytics.

\subsubsection{Limitations Associated with ML Techniques}
This subsection outlines three relevant limitations that pertain to the ML techniques used.

\subsubsection*{Oversimplification of AHR data} 

Many of the AHR-based studies outlined explore simple relationships in AHRs, such as the similarity between drugs or diseases based on co-occurrence. The techniques used in these studies cannot capture complex health service patterns since the techniques used require simplification of temporal, highly-dimensional AHR data. When AHRs are simplified, much of their subtlety is lost, which precludes extraction of complex healthcare patterns and medical trajectories and pathways. Many studies acknowledge this issue, and seek to develop novel modelling techniques instead of simplifying the data. This trend is most evident with neural network-based techniques (Section~\ref{subsec:network_analysis}), where the majority of studies use novel methods to capture such complex relationships, often through representation learning.

\subsubsection*{Lack of Population-scale Studies} Another limitation is that many studies limit the scope of their methods to patient- or hospital-level, and do not attempt to analyse population-level patterns. This trend can often be attributed to AHR-related limitations, however it is also evident that such population-level studies require significant abstraction, filtering, and assumptions informed by domain knowledge. This unique challenge is also not addressed by any established techniques, and instead requires the development of new and novel solutions. 

\subsubsection*{Scalability} Memory requirements are a common limitation in ML analytics, particularly in network science and deep learning approaches. These methods often require a complex composition of data structures and algorithms, especially during model training. In many AHR-related studies, scalability issues are often mitigated by simplifying AHRs to the point that they lose much of the complex and temporal relationships inherent in the records. 

\section{Open Research Questions}\label{sec:openRQs}
This section discuss open research questions in the area of population-level health service pattern discovery we identified, based on the state-of-art literature and emerging research areas that  utilise AHRs.

\subsubsection*{How can benchmark and synthetic data be used to accelerate AHR-related research?}

Benchmark datasets are public datasets, and thus do not exhibit many of the limitations associated with AHR ata, such as data accessibility, data quantity, and fragmentation. Benchmark datasets also make AHR-related research accessible to more researchers. Many of the studies explored in this review employ benchmark datasets for evaluating their methods. Despite widespread use of benchmark datasets, there are not many available to the public for research. The most commonly used benchmark datasets include the MIMIC-III \cite{johnson2016mimic} and -IV \cite{johnson2020mimic} (containing intensive care unit AHRs), and the BPIC-11 \cite{vanDongen2011Dutch} (Gynaecology department AHRs). Benchmark datasets are critical to the development of novel modelling techniques, for evaluating and comparing these techniques with a wider audience, and  for further progressing AHR-based research. Availability of many diverse benchmark datasets will help in these endeavours. 

Data synthesis methods can also be a useful way to share datasets and develop new modelling techniques \cite{Raghunathan_2021} as it can mitigate data privacy issues and data quality limitations. Many methods have been established specifically for synthesising patient-level data \cite{Goncalves_2020,Tucker_2020}. In comparison to benchmark data, synthetic data is not a widely used technique for mitigating the various AHR-related limitations. Synthetic data should be used to help accelerate development of research in this area.

\subsubsection*{How can AHRs be accurately represented and visualised?}

AHRs often contain indirect relationships to high-level healthcare concepts (as discussed in \ref{sec:ahr_char} and evidenced by the many indirect-analytics studies shown in Figure~\ref{fig:rq3}), which makes it challenging to explore the data to find these relationships. Many of the limitations associated with AHR-data -- such as data quality, fragmentation, and irregular temporal data -- also inhibit representation and visualisation efforts. This review highlighted many ways in which AHRs can be represented and visualised for domain-specific tasks, however many of these representations and visualisations thereof are abstract. New methods could be developed to make interpretable representations and visualisations of AHRs.

\subsubsection*{Can dimensionality reduction be applied to temporal AHR data?}

Some of the identified studies use dimensionality reduction as a data pre-processing or post-processing step, generating features for downstream analytics. However, dimensionality reduction can also be a useful method for data exploration, but it was only used in one study~\cite{chen2019mining}. One reason for this could be the irregular temporal nature of AHR-data; while some methods have been developed to reduce the dimension of temporal data \cite{gashler2011temporal,ali2019timecluster,lewandowski2010temporal}, we did not observe any applied to AHRs.

\subsubsection*{Can high-level healthcare concepts be modelled as latent variables?}

The indirect nature of AHR data was a limitation recognised by some of the studies explored in this review. These studies seek to model the unobserved, latent variables present in AHRs using a variety of techniques, such as Hidden Markov Models (HMMs) \cite{huang2018probabilistic,leontjeva2016complex,Bueno_2018} and topic models \cite{prokofyeva2019application,Huang_2015,Huang_2015a}. Future research should examine how these latent variables relate to various high-level concepts such as analysis of complex healthcare patterns or medical trajectories and pathways since these healthcare concepts are not recorded in AHRs.

\subsubsection*{How can AHR analytics insights be validated and interpreted?}

Clinical data mining models must be validated to a standard that is much higher than in many other fields. Typically, data mining validation commonly used other fields (such as digital marketing or retail) seeks to demonstrate reproducible performance using standard techniques such as cross-validation. However, validity in healthcare must be applied at a cohort-level. 

With recent advances in machine learning (e.g., deep learning), understanding a model's decision process is becoming increasingly difficult. Healthcare is a field where the interpretability of a model's decision process is critical. There are recent advancements in ExplainableAI, with recent studies such as \cite{liu2022interpretable} that use Shapley value based interpretations, proposing workarounds for black-box nature of state. Interpretability leads to trustworthiness \cite{lipton2018mythos} and constructing trustworthy models is advantageous for the adoption of machine-learned models in clinical practice. Novel methods are needed to validate and interpret ML models.

\subsubsection*{How can the interoperability of AHRs be maximised?}

Health standards, vocabulary, and practices vary across different countries and continents. Achieving true data interoperability requires the development and implementation of standards and clinical-content models for the unambiguous representation and exchange of clinical meaning. Various international certification and standards bodies pursue this goal and maintain resources such as international clinical terminology (CT)\footnote{SNOMED clinical terminology from the International Health Terminology Standards Development Organisation (IHTSDO) http://www.ihtsdo.org} and International Disease Classification (ICD)\footnote{ICD codes are standardised by the World Health Organisation (WHO) http://www.who.int/classifications/icd/en} \cite{ICD_10}. AHR-based research should pursue solutions that are interoperable across these different standards. AHR researchers and data custodians should apply the FAIR data principles \cite{wilkinson2016fair} in pursuit of this goal.

\xspace 

\section{Discussion}\label{sec:discussion} 
 
This study highlights the trends in AHR-based research. Clustering, neural networks, and network analysis were among the most widely applied techniques used for analysing AHRs. 

The most common application of AHR-based research is for understanding and predicting patient's medical trajectories. Other common applications include the construction of evidence-based guidelines, analysing patient comorbitites or cohorts and for biomarker discovery. 

A common challenge for the identified ML methods is the use of abstract features, that make it difficult to explore the complex features within AHRs. Neural networks were the most widely applied technique to overcome this challenge, seeking to learn informative, encoded representations of AHRs. Neural networks struggle however in generating easily interpretable results, with learned representations requiring further downstream processing and analytics. Other studies address the abstract feature issue by modelling latent variables in AHRs with Markov-based techniques, topic modelling, or Bayesian networks. These studies typically use patient features, such as health state or treatment behaviours, to understand clinical pathways. 

While this systematic review focuses specifically on AHRs, adjacent areas of research that use AHRs, such as the use of AHRs in combination with other EHR modalities or with genomic data, are not within the scope of this study. Using a diverse set of data modalities could contribute to a better understanding of health service patterns, but may also require novel analytics techniques and methods for combining knowledge from differing data modalities.

\section{Conclusion}\label{sec:conclusion}

In this study, we identify the significance of AHRs in emerging health informatics research, and provide a considered definition for the term. AHRs are principally used to pursue health service pattern mining applications, particularly for learning about population-health.

This systematic review answers four key research questions concerning the use of AHRs for HSPM. We identify twelve machine learning techniques that are used to analyse AHRs, and eight key health informatics applications. Furthermore, we analyse how each technique is applied and which techniques are used in pursuit of each application, revealing the current landscape of AHR-based research. Finally, this analysis highlighted many data-related and technique-related limitations of AHR-based research. We also identify six open research questions to provide clarity on the current state of AHR-based research and future research opportunities in this area. 

It is clear that the use of AHRs in health informatics research is becoming more commonplace among researchers and healthcare institutions such as hospitals and governments. At the same time, machine learning techniques are advancing and are consequently being used to pursue increasingly complex research objectives. These trends are reaffirmed by the analysis in this review, with emerging machine learning techniques being used to analyse latent and complex healthcare patterns in AHRs.


\section*{Acknowledgements}

This project was supported by a Commonwealth Services Contract (CA-DATA-200421) funded by Cancer Australia. 

The authors thank Massimo Piccardi for providing feedback on a draft of this manuscript.

%


\bibliographystyle{elsarticle-num-names}

\bibliography{mybibfile,ref}

\begin{thebibliography}{112}
\expandafter\ifx\csname natexlab\endcsname\relax\def\natexlab#1{#1}\fi
\providecommand{\url}[1]{\texttt{#1}}
\providecommand{\href}[2]{#2}
\providecommand{\path}[1]{#1}
\providecommand{\DOIprefix}{doi:}
\providecommand{\ArXivprefix}{arXiv:}
\providecommand{\URLprefix}{URL: }
\providecommand{\Pubmedprefix}{pmid:}
\providecommand{\doi}[1]{\href{http://dx.doi.org/#1}{\path{#1}}}
\providecommand{\Pubmed}[1]{\href{pmid:#1}{\path{#1}}}
\providecommand{\bibinfo}[2]{#2}
\ifx\xfnm\relax \def\xfnm[#1]{\unskip,\space#1}\fi
\bibitem[{Pramanik et~al.(2019)Pramanik, Pal, and
  Mukhopadhyay}]{pramanik2019healthcare}
\bibinfo{author}{P.~K.~D. Pramanik}, \bibinfo{author}{S.~Pal},
  \bibinfo{author}{M.~Mukhopadhyay},
\newblock \bibinfo{title}{Healthcare big data: A comprehensive overview},
\newblock \bibinfo{journal}{Intelligent systems for healthcare management and
  delivery}  (\bibinfo{year}{2019}) \bibinfo{pages}{72--100}.
\bibitem[{Shah and Khan(2020)}]{shah2020secondary}
\bibinfo{author}{S.~M. Shah}, \bibinfo{author}{R.~A. Khan},
\newblock \bibinfo{title}{Secondary use of electronic health record:
  Opportunities and challenges},
\newblock \bibinfo{journal}{IEEE Access} \bibinfo{volume}{8}
  (\bibinfo{year}{2020}) \bibinfo{pages}{136947--136965}.
\bibitem[{Yadav et~al.(2018)Yadav, Steinbach, Kumar, and
  Simon}]{yadav2018mining}
\bibinfo{author}{P.~Yadav}, \bibinfo{author}{M.~Steinbach},
  \bibinfo{author}{V.~Kumar}, \bibinfo{author}{G.~Simon},
\newblock \bibinfo{title}{Mining electronic health records ({EHR}s) a survey},
\newblock \bibinfo{journal}{ACM Computing Surveys (CSUR)} \bibinfo{volume}{50}
  (\bibinfo{year}{2018}) \bibinfo{pages}{1--40}.
\bibitem[{Chen et~al.(2017)Chen, Wei, Guo, Tang, and Sun}]{chen2017textual}
\bibinfo{author}{J.~Chen}, \bibinfo{author}{W.~Wei}, \bibinfo{author}{C.~Guo},
  \bibinfo{author}{L.~Tang}, \bibinfo{author}{L.~Sun},
\newblock \bibinfo{title}{Textual analysis and visualization of research trends
  in data mining for electronic health records},
\newblock \bibinfo{journal}{Health Policy and Technology} \bibinfo{volume}{6}
  (\bibinfo{year}{2017}) \bibinfo{pages}{389--400}.
\bibitem[{Kurniati et~al.(2016)Kurniati, Johnson, Hogg, and
  Hall}]{kurniati2016process}
\bibinfo{author}{A.~P. Kurniati}, \bibinfo{author}{O.~Johnson},
  \bibinfo{author}{D.~Hogg}, \bibinfo{author}{G.~Hall},
\newblock \bibinfo{title}{Process mining in oncology: A literature review},
\newblock in: \bibinfo{booktitle}{2016 6th International Conference on
  Information Communication and Management (ICICM)},
  \bibinfo{organization}{IEEE}, \bibinfo{year}{2016}, pp.
  \bibinfo{pages}{291--297}.
\bibitem[{Rojas et~al.(2016)Rojas, Munoz-Gama, Sep{\'u}lveda, and
  Capurro}]{rojas2016process}
\bibinfo{author}{E.~Rojas}, \bibinfo{author}{J.~Munoz-Gama},
  \bibinfo{author}{M.~Sep{\'u}lveda}, \bibinfo{author}{D.~Capurro},
\newblock \bibinfo{title}{Process mining in healthcare: A literature review},
\newblock \bibinfo{journal}{Journal of Biomedical Informatics}
  \bibinfo{volume}{61} (\bibinfo{year}{2016}) \bibinfo{pages}{224--236}.
\bibitem[{Erdogan and Tarhan(2018)}]{erdogan2018systematic}
\bibinfo{author}{T.~G. Erdogan}, \bibinfo{author}{A.~Tarhan},
\newblock \bibinfo{title}{Systematic mapping of process mining studies in
  healthcare},
\newblock \bibinfo{journal}{IEEE Access} \bibinfo{volume}{6}
  (\bibinfo{year}{2018}) \bibinfo{pages}{24543--24567}.
\bibitem[{Guzzo et~al.(2022)Guzzo, Rullo, and Vocaturo}]{guzzo2022process}
\bibinfo{author}{A.~Guzzo}, \bibinfo{author}{A.~Rullo},
  \bibinfo{author}{E.~Vocaturo},
\newblock \bibinfo{title}{Process mining applications in the healthcare domain:
  A comprehensive review},
\newblock \bibinfo{journal}{Wiley Interdisciplinary Reviews: Data Mining and
  Knowledge Discovery} \bibinfo{volume}{12} (\bibinfo{year}{2022})
  \bibinfo{pages}{e1442}.
\bibitem[{Munoz-Gama et~al.(2022)Munoz-Gama, Martin, Fernandez-Llatas, Johnson,
  Sep{\'u}lveda, Helm, Galvez-Yanjari, Rojas, Martinez-Millana, Aloini
  et~al.}]{munoz2022process}
\bibinfo{author}{J.~Munoz-Gama}, \bibinfo{author}{N.~Martin},
  \bibinfo{author}{C.~Fernandez-Llatas}, \bibinfo{author}{O.~A. Johnson},
  \bibinfo{author}{M.~Sep{\'u}lveda}, \bibinfo{author}{E.~Helm},
  \bibinfo{author}{V.~Galvez-Yanjari}, \bibinfo{author}{E.~Rojas},
  \bibinfo{author}{A.~Martinez-Millana}, \bibinfo{author}{D.~Aloini}, et~al.,
\newblock \bibinfo{title}{Process mining for healthcare: Characteristics and
  challenges},
\newblock \bibinfo{journal}{Journal of Biomedical Informatics}
  \bibinfo{volume}{127} (\bibinfo{year}{2022}) \bibinfo{pages}{103994}.
\bibitem[{Brunson and Laubenbacher(2017)}]{Brunson_2017}
\bibinfo{author}{J.~C. Brunson}, \bibinfo{author}{R.~C. Laubenbacher},
\newblock \bibinfo{title}{Applications of network analysis to routinely
  collected health care data: a systematic review},
\newblock \bibinfo{journal}{Journal of the American Medical Informatics
  Association} \bibinfo{volume}{25} (\bibinfo{year}{2017})
  \bibinfo{pages}{210--221}. \URLprefix
  \url{https://doi.org/10.1093%2Fjamia%2Focx052}.
  \DOIprefix\doi{10.1093/jamia/ocx052}.
\bibitem[{Shickel et~al.(2017)Shickel, Tighe, Bihorac, and
  Rashidi}]{shickel2017deep}
\bibinfo{author}{B.~Shickel}, \bibinfo{author}{P.~J. Tighe},
  \bibinfo{author}{A.~Bihorac}, \bibinfo{author}{P.~Rashidi},
\newblock \bibinfo{title}{Deep {EHR}: a survey of recent advances in deep
  learning techniques for electronic health record ({EHR}) analysis},
\newblock \bibinfo{journal}{IEEE Journal of Biomedical and Health Informatics}
  \bibinfo{volume}{22} (\bibinfo{year}{2017}) \bibinfo{pages}{1589--1604}.
\bibitem[{Xiao et~al.(2018)Xiao, Choi, and Sun}]{xiao2018opportunities}
\bibinfo{author}{C.~Xiao}, \bibinfo{author}{E.~Choi}, \bibinfo{author}{J.~Sun},
\newblock \bibinfo{title}{Opportunities and challenges in developing deep
  learning models using electronic health records data: a systematic review},
\newblock \bibinfo{journal}{Journal of the American Medical Informatics
  Association} \bibinfo{volume}{25} (\bibinfo{year}{2018})
  \bibinfo{pages}{1419--1428}.
\bibitem[{Cadarette and Wong(2015)}]{cadarette2015introduction}
\bibinfo{author}{S.~M. Cadarette}, \bibinfo{author}{L.~Wong},
\newblock \bibinfo{title}{An introduction to health care administrative data},
\newblock \bibinfo{journal}{The Canadian Journal of Hospital Pharmacy}
  \bibinfo{volume}{68} (\bibinfo{year}{2015}) \bibinfo{pages}{232}.
\bibitem[{{Australian Institute of Health and Wellness
  (AIHW)}(2021)}]{AIHW_AHR}
\bibinfo{author}{{Australian Institute of Health and Wellness (AIHW)}},
  \bibinfo{title}{Our data collections},
  \bibinfo{howpublished}{\url{https://www.aihw.gov.au/about-our-data/our-data-collections}},
  \bibinfo{year}{2021}. \bibinfo{note}{Accessed: 2022-04-11}.
\bibitem[{{National Health Service (NHS)}(2021)}]{NHS_AHR}
\bibinfo{author}{{National Health Service (NHS)}}, \bibinfo{title}{List of
  administrative sources},
  \bibinfo{howpublished}{\url{https://digital.nhs.uk/data-and-information/find-data-and-publications/statement-of-administrative-sources/list-of-administrative-sources}},
  \bibinfo{year}{2021}. \bibinfo{note}{Accessed: 2022-04-11}.
\bibitem[{Kindig and Stoddart(2003)}]{kindig2003population}
\bibinfo{author}{D.~Kindig}, \bibinfo{author}{G.~Stoddart},
\newblock \bibinfo{title}{What is population health?},
\newblock \bibinfo{journal}{American journal of public health}
  \bibinfo{volume}{93} (\bibinfo{year}{2003}) \bibinfo{pages}{380--383}.
\bibitem[{Rebuge and Ferreira(2012)}]{rebuge2012business}
\bibinfo{author}{{\'A}.~Rebuge}, \bibinfo{author}{D.~R. Ferreira},
\newblock \bibinfo{title}{Business process analysis in healthcare environments:
  A methodology based on process mining},
\newblock \bibinfo{journal}{Information systems} \bibinfo{volume}{37}
  (\bibinfo{year}{2012}) \bibinfo{pages}{99--116}.
\bibitem[{Newman(2018)}]{newman2018networks}
\bibinfo{author}{M.~Newman}, \bibinfo{title}{Networks},
  \bibinfo{publisher}{Oxford University Press}, \bibinfo{year}{2018}.
\bibitem[{Petersen et~al.(2008)Petersen, Feldt, Mujtaba, and
  Mattsson}]{petersen2008systematic}
\bibinfo{author}{K.~Petersen}, \bibinfo{author}{R.~Feldt},
  \bibinfo{author}{S.~Mujtaba}, \bibinfo{author}{M.~Mattsson},
\newblock \bibinfo{title}{Systematic mapping studies in software engineering},
\newblock in: \bibinfo{booktitle}{12th International Conference on Evaluation
  and Assessment in Software Engineering (EASE) 12}, \bibinfo{year}{2008}, pp.
  \bibinfo{pages}{1--10}.
\bibitem[{Harris et~al.(2014)Harris, Quatman, Manring, Siston, and
  Flanigan}]{harris2014write}
\bibinfo{author}{J.~D. Harris}, \bibinfo{author}{C.~E. Quatman},
  \bibinfo{author}{M.~Manring}, \bibinfo{author}{R.~A. Siston},
  \bibinfo{author}{D.~C. Flanigan},
\newblock \bibinfo{title}{How to write a systematic review},
\newblock \bibinfo{journal}{The American journal of sports medicine}
  \bibinfo{volume}{42} (\bibinfo{year}{2014}) \bibinfo{pages}{2761--2768}.
\bibitem[{Wohlin(2014)}]{wohlin2014guidelines}
\bibinfo{author}{C.~Wohlin},
\newblock \bibinfo{title}{Guidelines for snowballing in systematic literature
  studies and a replication in software engineering},
\newblock in: \bibinfo{booktitle}{Proceedings of the 18th international
  conference on evaluation and assessment in software engineering},
  \bibinfo{year}{2014}, pp. \bibinfo{pages}{1--10}.
\bibitem[{Page et~al.(2021)Page, McKenzie, Bossuyt, Boutron, Hoffmann, Mulrow,
  Shamseer, Tetzlaff, Akl, Brennan, Chou, Glanville, Grimshaw,
  Hr{\'o}bjartsson, Lalu, Li, Loder, Mayo-Wilson, McDonald, McGuinness,
  Stewart, Thomas, Tricco, Welch, Whiting, and Moher}]{Pagen71}
\bibinfo{author}{M.~J. Page}, \bibinfo{author}{J.~E. McKenzie},
  \bibinfo{author}{P.~M. Bossuyt}, \bibinfo{author}{I.~Boutron},
  \bibinfo{author}{T.~C. Hoffmann}, \bibinfo{author}{C.~D. Mulrow},
  \bibinfo{author}{L.~Shamseer}, \bibinfo{author}{J.~M. Tetzlaff},
  \bibinfo{author}{E.~A. Akl}, \bibinfo{author}{S.~E. Brennan},
  \bibinfo{author}{R.~Chou}, \bibinfo{author}{J.~Glanville},
  \bibinfo{author}{J.~M. Grimshaw}, \bibinfo{author}{A.~Hr{\'o}bjartsson},
  \bibinfo{author}{M.~M. Lalu}, \bibinfo{author}{T.~Li}, \bibinfo{author}{E.~W.
  Loder}, \bibinfo{author}{E.~Mayo-Wilson}, \bibinfo{author}{S.~McDonald},
  \bibinfo{author}{L.~A. McGuinness}, \bibinfo{author}{L.~A. Stewart},
  \bibinfo{author}{J.~Thomas}, \bibinfo{author}{A.~C. Tricco},
  \bibinfo{author}{V.~A. Welch}, \bibinfo{author}{P.~Whiting},
  \bibinfo{author}{D.~Moher},
\newblock \bibinfo{title}{The {PRISMA} 2020 statement: an updated guideline for
  reporting systematic reviews},
\newblock \bibinfo{journal}{British Medical Journal} \bibinfo{volume}{372}
  (\bibinfo{year}{2021}). \URLprefix
  \url{https://www.bmj.com/content/372/bmj.n71}.
  \DOIprefix\doi{10.1136/bmj.n71}.
  \href{http://arxiv.org/abs/https://www.bmj.com/content/372/bmj.n71.full.pdf}{{\tt
  arXiv:https://www.bmj.com/content/372/bmj.n71.full.pdf}}.
\bibitem[{Khan et~al.(2011)Khan, Kunz, Kleijnen, and
  Antes}]{khan2011systematic}
\bibinfo{author}{K.~Khan}, \bibinfo{author}{R.~Kunz},
  \bibinfo{author}{J.~Kleijnen}, \bibinfo{author}{G.~Antes},
  \bibinfo{title}{Systematic reviews to support evidence-based medicine},
  \bibinfo{publisher}{CRC press}, \bibinfo{year}{2011}.
\bibitem[{Choi et~al.(2016)Choi, Bahadori, Searles, Coffey, Thompson, Bost,
  Tejedor-Sojo, and Sun}]{choi2016multi}
\bibinfo{author}{E.~Choi}, \bibinfo{author}{M.~T. Bahadori},
  \bibinfo{author}{E.~Searles}, \bibinfo{author}{C.~Coffey},
  \bibinfo{author}{M.~Thompson}, \bibinfo{author}{J.~Bost},
  \bibinfo{author}{J.~Tejedor-Sojo}, \bibinfo{author}{J.~Sun},
\newblock \bibinfo{title}{Multi-layer representation learning for medical
  concepts},
\newblock in: \bibinfo{booktitle}{proceedings of the 22nd ACM SIGKDD
  international conference on knowledge discovery and data mining},
  \bibinfo{year}{2016}, pp. \bibinfo{pages}{1495--1504}.
\bibitem[{Huang et~al.(2015)Huang, Dong, Bath, Ji, and Duan}]{huang2015mining}
\bibinfo{author}{Z.~Huang}, \bibinfo{author}{W.~Dong},
  \bibinfo{author}{P.~Bath}, \bibinfo{author}{L.~Ji},
  \bibinfo{author}{H.~Duan},
\newblock \bibinfo{title}{On mining latent treatment patterns from electronic
  medical records},
\newblock \bibinfo{journal}{Data mining and knowledge discovery}
  \bibinfo{volume}{29} (\bibinfo{year}{2015}) \bibinfo{pages}{914--949}.
\bibitem[{Jensen et~al.(2014)Jensen, Moseley, Oprea, Elles{\o}e, Eriksson,
  Schmock, Jensen, Jensen, and Brunak}]{jensen2014temporal}
\bibinfo{author}{A.~B. Jensen}, \bibinfo{author}{P.~L. Moseley},
  \bibinfo{author}{T.~I. Oprea}, \bibinfo{author}{S.~G. Elles{\o}e},
  \bibinfo{author}{R.~Eriksson}, \bibinfo{author}{H.~Schmock},
  \bibinfo{author}{P.~B. Jensen}, \bibinfo{author}{L.~J. Jensen},
  \bibinfo{author}{S.~Brunak},
\newblock \bibinfo{title}{Temporal disease trajectories condensed from
  population-wide registry data covering 6.2 million patients},
\newblock \bibinfo{journal}{Nature communications} \bibinfo{volume}{5}
  (\bibinfo{year}{2014}) \bibinfo{pages}{1--10}.
\bibitem[{Liu et~al.(2015)Liu, Wang, Hu, and Xiong}]{liu2015temporal}
\bibinfo{author}{C.~Liu}, \bibinfo{author}{F.~Wang}, \bibinfo{author}{J.~Hu},
  \bibinfo{author}{H.~Xiong},
\newblock \bibinfo{title}{Temporal phenotyping from longitudinal electronic
  health records: A graph based framework},
\newblock in: \bibinfo{booktitle}{Proceedings of the 21th ACM SIGKDD
  international conference on knowledge discovery and data mining},
  \bibinfo{year}{2015}, pp. \bibinfo{pages}{705--714}.
\bibitem[{Baker et~al.(2017)Baker, Dunwoodie, Jones, Newsham, Johnson, Price,
  Wolstenholme, Leal, McGinley, Twelves et~al.}]{baker2017process}
\bibinfo{author}{K.~Baker}, \bibinfo{author}{E.~Dunwoodie},
  \bibinfo{author}{R.~G. Jones}, \bibinfo{author}{A.~Newsham},
  \bibinfo{author}{O.~Johnson}, \bibinfo{author}{C.~P. Price},
  \bibinfo{author}{J.~Wolstenholme}, \bibinfo{author}{J.~Leal},
  \bibinfo{author}{P.~McGinley}, \bibinfo{author}{C.~Twelves}, et~al.,
\newblock \bibinfo{title}{Process mining routinely collected electronic health
  records to define real-life clinical pathways during chemotherapy},
\newblock \bibinfo{journal}{International journal of medical informatics}
  \bibinfo{volume}{103} (\bibinfo{year}{2017}) \bibinfo{pages}{32--41}.
\bibitem[{Choi et~al.(2017)Choi, Bahadori, Song, Stewart, and
  Sun}]{choi2017gram}
\bibinfo{author}{E.~Choi}, \bibinfo{author}{M.~T. Bahadori},
  \bibinfo{author}{L.~Song}, \bibinfo{author}{W.~F. Stewart},
  \bibinfo{author}{J.~Sun},
\newblock \bibinfo{title}{{GRAM}: graph-based attention model for healthcare
  representation learning},
\newblock in: \bibinfo{booktitle}{Proceedings of the 23rd ACM SIGKDD
  international conference on knowledge discovery and data mining},
  \bibinfo{year}{2017}, pp. \bibinfo{pages}{787--795}.
\bibitem[{Choi et~al.(2018)Choi, Xiao, Stewart, and Sun}]{choi2018mime}
\bibinfo{author}{E.~Choi}, \bibinfo{author}{C.~Xiao}, \bibinfo{author}{W.~F.
  Stewart}, \bibinfo{author}{J.~Sun},
\newblock \bibinfo{title}{{MiME}: multilevel medical embedding of electronic
  health records for predictive healthcare},
\newblock in: \bibinfo{booktitle}{Proceedings of the 32nd International
  Conference on Neural Information Processing Systems}, \bibinfo{year}{2018},
  pp. \bibinfo{pages}{4552--4562}.
\bibitem[{Guo et~al.(2020)Guo, Fujiwara, Li, Lima, Sen, Tran, and
  Ma}]{guo2020comparative}
\bibinfo{author}{R.~Guo}, \bibinfo{author}{T.~Fujiwara},
  \bibinfo{author}{Y.~Li}, \bibinfo{author}{K.~M. Lima},
  \bibinfo{author}{S.~Sen}, \bibinfo{author}{N.~K. Tran},
  \bibinfo{author}{K.-L. Ma},
\newblock \bibinfo{title}{Comparative visual analytics for assessing medical
  records with sequence embedding},
\newblock \bibinfo{journal}{Visual Informatics} \bibinfo{volume}{4}
  (\bibinfo{year}{2020}) \bibinfo{pages}{72--85}.
\bibitem[{Pham et~al.(2017)Pham, Tran, Phung, and
  Venkatesh}]{pham2017predicting}
\bibinfo{author}{T.~Pham}, \bibinfo{author}{T.~Tran},
  \bibinfo{author}{D.~Phung}, \bibinfo{author}{S.~Venkatesh},
\newblock \bibinfo{title}{Predicting healthcare trajectories from medical
  records: A deep learning approach},
\newblock \bibinfo{journal}{Journal of biomedical informatics}
  \bibinfo{volume}{69} (\bibinfo{year}{2017}) \bibinfo{pages}{218--229}.
\bibitem[{Zhang et~al.(2018)Zhang, Kowsari, Harrison, Lobo, and
  Barnes}]{zhang2018patient2vec}
\bibinfo{author}{J.~Zhang}, \bibinfo{author}{K.~Kowsari},
  \bibinfo{author}{J.~H. Harrison}, \bibinfo{author}{J.~M. Lobo},
  \bibinfo{author}{L.~E. Barnes},
\newblock \bibinfo{title}{Patient2vec: A personalized interpretable deep
  representation of the longitudinal electronic health record},
\newblock \bibinfo{journal}{IEEE Access} \bibinfo{volume}{6}
  (\bibinfo{year}{2018}) \bibinfo{pages}{65333--65346}.
\bibitem[{Jin et~al.(2018)Jin, Yang, Sun, Liu, Qu, and Tong}]{jin2018treatment}
\bibinfo{author}{B.~Jin}, \bibinfo{author}{H.~Yang}, \bibinfo{author}{L.~Sun},
  \bibinfo{author}{C.~Liu}, \bibinfo{author}{Y.~Qu}, \bibinfo{author}{J.~Tong},
\newblock \bibinfo{title}{A treatment engine by predicting next-period
  prescriptions},
\newblock in: \bibinfo{booktitle}{Proceedings of the 24th ACM SIGKDD
  International Conference on Knowledge Discovery \& Data Mining},
  \bibinfo{year}{2018}, pp. \bibinfo{pages}{1608--1616}.
\bibitem[{Li et~al.(2020)Li, Zhu, Wu, Ding, and Zhao}]{li2020ccae}
\bibinfo{author}{Y.~Li}, \bibinfo{author}{Z.~Zhu}, \bibinfo{author}{H.~Wu},
  \bibinfo{author}{S.~Ding}, \bibinfo{author}{Y.~Zhao},
\newblock \bibinfo{title}{{CCAE}: Cross-field categorical attributes embedding
  for cancer clinical endpoint prediction},
\newblock \bibinfo{journal}{Artificial Intelligence in Medicine}
  \bibinfo{volume}{107} (\bibinfo{year}{2020}) \bibinfo{pages}{101915}.
\bibitem[{Xu et~al.(2018)Xu, Wang, Jin, and Wang}]{xu2018learning}
\bibinfo{author}{X.~Xu}, \bibinfo{author}{Y.~Wang}, \bibinfo{author}{T.~Jin},
  \bibinfo{author}{J.~Wang},
\newblock \bibinfo{title}{Learning the representation of medical features for
  clinical pathway analysis},
\newblock in: \bibinfo{booktitle}{International Conference on Database Systems
  for Advanced Applications}, \bibinfo{organization}{Springer},
  \bibinfo{year}{2018}, pp. \bibinfo{pages}{37--52}.
\bibitem[{Beaulieu-Jones et~al.(2018)Beaulieu-Jones, Orzechowski, and
  Moore}]{beaulieu2018mapping}
\bibinfo{author}{B.~K. Beaulieu-Jones}, \bibinfo{author}{P.~Orzechowski},
  \bibinfo{author}{J.~H. Moore},
\newblock \bibinfo{title}{Mapping patient trajectories using longitudinal
  extraction and deep learning in the {MIMIC-III} critical care database},
\newblock in: \bibinfo{booktitle}{Pacific Symposium on Biocomputing 2018:
  Proceedings of the Pacific Symposium}, \bibinfo{organization}{World
  Scientific}, \bibinfo{year}{2018}, pp. \bibinfo{pages}{123--132}.
\bibitem[{Hong et~al.(2017)Hong, Wu, Li, and Wu}]{Hong_2017}
\bibinfo{author}{S.~Hong}, \bibinfo{author}{M.~Wu}, \bibinfo{author}{H.~Li},
  \bibinfo{author}{Z.~Wu},
\newblock \bibinfo{title}{Event2{V}ec: Learning representations of events on
  temporal sequences},
\newblock in: \bibinfo{booktitle}{Web and Big Data},
  \bibinfo{publisher}{Springer International Publishing}, \bibinfo{year}{2017},
  pp. \bibinfo{pages}{33--47}. \URLprefix
  \url{https://doi.org/10.1007%2F978-3-319-63564-4_3}.
  \DOIprefix\doi{10.1007/978-3-319-63564-4_3}.
\bibitem[{Li et~al.(2020)Li, Zuo, Coston, Weiss, and Chen}]{Li_2020}
\bibinfo{author}{L.~Li}, \bibinfo{author}{R.~Zuo}, \bibinfo{author}{A.~Coston},
  \bibinfo{author}{J.~C. Weiss}, \bibinfo{author}{G.~H. Chen},
\newblock \bibinfo{title}{Neural topic models with survival supervision:
  Jointly predicting time-to-event outcomes and learning how clinical features
  relate},
\newblock in: \bibinfo{booktitle}{Artificial Intelligence in Medicine},
  \bibinfo{publisher}{Springer International Publishing}, \bibinfo{year}{2020},
  pp. \bibinfo{pages}{371--381}. \URLprefix
  \url{https://doi.org/10.1007%2F978-3-030-59137-3_33}.
  \DOIprefix\doi{10.1007/978-3-030-59137-3_33}.
\bibitem[{Lu et~al.(2021)Lu, Cui, Sun, and Zhu}]{Lu_2021}
\bibinfo{author}{X.~Lu}, \bibinfo{author}{L.~Cui}, \bibinfo{author}{Z.~Sun},
  \bibinfo{author}{Y.~Zhu},
\newblock \bibinfo{title}{{ProAID}: path-based reasoning for self-attentional
  disease prediction},
\newblock \bibinfo{journal}{Knowledge and Information Systems}
  \bibinfo{volume}{63} (\bibinfo{year}{2021}) \bibinfo{pages}{3087--3101}.
  \URLprefix \url{https://doi.org/10.1007%2Fs10115-021-01617-w}.
  \DOIprefix\doi{10.1007/s10115-021-01617-w}.
\bibitem[{Wolff et~al.(2020)Wolff, Gary, Jung, Normann, Kaier, Binder,
  Domschke, Klimke, and Franz}]{Wolff_2020}
\bibinfo{author}{J.~Wolff}, \bibinfo{author}{A.~Gary},
  \bibinfo{author}{D.~Jung}, \bibinfo{author}{C.~Normann},
  \bibinfo{author}{K.~Kaier}, \bibinfo{author}{H.~Binder},
  \bibinfo{author}{K.~Domschke}, \bibinfo{author}{A.~Klimke},
  \bibinfo{author}{M.~Franz},
\newblock \bibinfo{title}{Predicting patient outcomes in psychiatric hospitals
  with routine data: a machine learning approach},
\newblock \bibinfo{journal}{{BMC} Medical Informatics and Decision Making}
  \bibinfo{volume}{20} (\bibinfo{year}{2020}) \bibinfo{pages}{1--9}. \URLprefix
  \url{https://doi.org/10.1186%2Fs12911-020-1042-2}.
  \DOIprefix\doi{10.1186/s12911-020-1042-2}.
\bibitem[{Steinberg et~al.(2021)Steinberg, Jung, Fries, Corbin, Pfohl, and
  Shah}]{steinberg2021language}
\bibinfo{author}{E.~Steinberg}, \bibinfo{author}{K.~Jung},
  \bibinfo{author}{J.~A. Fries}, \bibinfo{author}{C.~K. Corbin},
  \bibinfo{author}{S.~R. Pfohl}, \bibinfo{author}{N.~H. Shah},
\newblock \bibinfo{title}{Language models are an effective representation
  learning technique for electronic health record data},
\newblock \bibinfo{journal}{Journal of Biomedical Informatics}
  \bibinfo{volume}{113} (\bibinfo{year}{2021}) \bibinfo{pages}{103637}.
\bibitem[{Zheng et~al.(2021)Zheng, Ryzhov, Xie, and Zhong}]{Zheng_2021}
\bibinfo{author}{H.~Zheng}, \bibinfo{author}{I.~O. Ryzhov},
  \bibinfo{author}{W.~Xie}, \bibinfo{author}{J.~Zhong},
\newblock \bibinfo{title}{Personalized multimorbidity management for patients
  with type 2 diabetes using reinforcement learning of electronic health
  records},
\newblock \bibinfo{journal}{Drugs} \bibinfo{volume}{81} (\bibinfo{year}{2021})
  \bibinfo{pages}{471--482}. \URLprefix
  \url{https://doi.org/10.1007%2Fs40265-020-01435-4}.
  \DOIprefix\doi{10.1007/s40265-020-01435-4}.
\bibitem[{Caruana et~al.(2022)Caruana, Bandara, Catchpoole, and
  Kennedy}]{Caruana_2022}
\bibinfo{author}{A.~Caruana}, \bibinfo{author}{M.~Bandara},
  \bibinfo{author}{D.~Catchpoole}, \bibinfo{author}{P.~J. Kennedy},
\newblock \bibinfo{title}{Beyond topics: Discovering latent healthcare
  objectives from event sequences},
\newblock in: \bibinfo{editor}{G.~Long}, \bibinfo{editor}{X.~Yu},
  \bibinfo{editor}{S.~Wang} (Eds.), \bibinfo{booktitle}{AI 2021: Advances in
  Artificial Intelligence}, \bibinfo{publisher}{Springer International
  Publishing}, \bibinfo{address}{Cham}, \bibinfo{year}{2022}, pp.
  \bibinfo{pages}{368--380}.
\bibitem[{Gerrard et~al.(2022)Gerrard, Peng, Clarke, Schlegel, and
  Jiang}]{Gerrard_2022}
\bibinfo{author}{L.~Gerrard}, \bibinfo{author}{X.~Peng},
  \bibinfo{author}{A.~Clarke}, \bibinfo{author}{C.~Schlegel},
  \bibinfo{author}{J.~Jiang},
\newblock \bibinfo{title}{Predicting outcomes for cancer patients with
  transformer-based multi-task learning},
\newblock in: \bibinfo{editor}{G.~Long}, \bibinfo{editor}{X.~Yu},
  \bibinfo{editor}{S.~Wang} (Eds.), \bibinfo{booktitle}{AI 2021: Advances in
  Artificial Intelligence}, \bibinfo{publisher}{Springer International
  Publishing}, \bibinfo{address}{Cham}, \bibinfo{year}{2022}, pp.
  \bibinfo{pages}{381--392}.
\bibitem[{Ochoa and Mustafa(2022)}]{ochoa2022graph}
\bibinfo{author}{J.~G.~D. Ochoa}, \bibinfo{author}{F.~E. Mustafa},
\newblock \bibinfo{title}{Graph neural network modelling as a potentially
  effective method for predicting and analyzing procedures based on patients'
  diagnoses},
\newblock \bibinfo{journal}{Artificial Intelligence in Medicine}
  \bibinfo{volume}{131} (\bibinfo{year}{2022}) \bibinfo{pages}{102359}.
\bibitem[{{Doshi-Velez} et~al.(2014){Doshi-Velez}, {Ge}, and
  {Kohane}}]{doshi-velez2014comorbidity}
\bibinfo{author}{F.~{Doshi-Velez}}, \bibinfo{author}{Y.~{Ge}},
  \bibinfo{author}{I.~{Kohane}},
\newblock \bibinfo{title}{Comorbidity clusters in autism spectrum disorders: An
  electronic health record time-series analysis},
\newblock \bibinfo{journal}{Pediatrics} \bibinfo{volume}{133}
  (\bibinfo{year}{2014}).
\bibitem[{Zhang et~al.(2015)Zhang, Padman, Wasserman, Patel, Teredesai, and
  Xie}]{Zhang_2015}
\bibinfo{author}{Y.~Zhang}, \bibinfo{author}{R.~Padman},
  \bibinfo{author}{L.~Wasserman}, \bibinfo{author}{N.~Patel},
  \bibinfo{author}{P.~Teredesai}, \bibinfo{author}{Q.~Xie},
\newblock \bibinfo{title}{On clinical pathway discovery from electronic health
  record data},
\newblock \bibinfo{journal}{{IEEE} Intelligent Systems} \bibinfo{volume}{30}
  (\bibinfo{year}{2015}) \bibinfo{pages}{70--75}. \URLprefix
  \url{https://doi.org/10.1109%2Fmis.2015.14}.
  \DOIprefix\doi{10.1109/mis.2015.14}.
\bibitem[{Roque et~al.(2011)Roque, Jensen, Schmock, Dalgaard, Andreatta,
  Hansen, S{\o}eby, Bredkj{\ae}r, Juul, Werge et~al.}]{roque2011using}
\bibinfo{author}{F.~S. Roque}, \bibinfo{author}{P.~B. Jensen},
  \bibinfo{author}{H.~Schmock}, \bibinfo{author}{M.~Dalgaard},
  \bibinfo{author}{M.~Andreatta}, \bibinfo{author}{T.~Hansen},
  \bibinfo{author}{K.~S{\o}eby}, \bibinfo{author}{S.~Bredkj{\ae}r},
  \bibinfo{author}{A.~Juul}, \bibinfo{author}{T.~Werge}, et~al.,
\newblock \bibinfo{title}{Using electronic patient records to discover disease
  correlations and stratify patient cohorts},
\newblock \bibinfo{journal}{PLOS Computational Biology} \bibinfo{volume}{7}
  (\bibinfo{year}{2011}) \bibinfo{pages}{e1002141}.
\bibitem[{Sideris et~al.(2016)Sideris, Pourhomayoun, Kalantarian, and
  Sarrafzadeh}]{sideris2016flexible}
\bibinfo{author}{C.~Sideris}, \bibinfo{author}{M.~Pourhomayoun},
  \bibinfo{author}{H.~Kalantarian}, \bibinfo{author}{M.~Sarrafzadeh},
\newblock \bibinfo{title}{A flexible data-driven comorbidity feature extraction
  framework},
\newblock \bibinfo{journal}{Computers in biology and medicine}
  \bibinfo{volume}{73} (\bibinfo{year}{2016}) \bibinfo{pages}{165--172}.
\bibitem[{Chen et~al.(2009)Chen, Blumm, Christakis, Barabasi, and
  Deisboeck}]{chen2009cancer}
\bibinfo{author}{L.~Chen}, \bibinfo{author}{N.~Blumm},
  \bibinfo{author}{N.~Christakis}, \bibinfo{author}{A.~Barabasi},
  \bibinfo{author}{T.~S. Deisboeck},
\newblock \bibinfo{title}{Cancer metastasis networks and the prediction of
  progression patterns},
\newblock \bibinfo{journal}{British journal of cancer} \bibinfo{volume}{101}
  (\bibinfo{year}{2009}) \bibinfo{pages}{749--758}.
\bibitem[{Chen et~al.(2018)Chen, Sun, Guo, Wei, and Xie}]{chen2018data}
\bibinfo{author}{J.~Chen}, \bibinfo{author}{L.~Sun}, \bibinfo{author}{C.~Guo},
  \bibinfo{author}{W.~Wei}, \bibinfo{author}{Y.~Xie},
\newblock \bibinfo{title}{A data-driven framework of typical treatment process
  extraction and evaluation},
\newblock \bibinfo{journal}{Journal of biomedical informatics}
  \bibinfo{volume}{83} (\bibinfo{year}{2018}) \bibinfo{pages}{178--195}.
\bibitem[{Apunike et~al.(2020)Apunike, Oliveira-Ciabati, Sanches, de~Oliveira,
  Sanchez, Galliez, and Alves}]{apunike2020analyses}
\bibinfo{author}{A.~C. Apunike}, \bibinfo{author}{L.~Oliveira-Ciabati},
  \bibinfo{author}{T.~L. Sanches}, \bibinfo{author}{L.~L. de~Oliveira},
  \bibinfo{author}{M.~N. Sanchez}, \bibinfo{author}{R.~M. Galliez},
  \bibinfo{author}{D.~Alves},
\newblock \bibinfo{title}{Analyses of public health databases via clinical
  pathway modelling: {TBWEB}},
\newblock in: \bibinfo{booktitle}{International Conference on Computational
  Science}, \bibinfo{organization}{Springer}, \bibinfo{year}{2020}, pp.
  \bibinfo{pages}{550--562}.
\bibitem[{Johns et~al.(2020)Johns, Hearne, Bernhardt, and
  Churilov}]{johns2020clustering}
\bibinfo{author}{H.~Johns}, \bibinfo{author}{J.~Hearne},
  \bibinfo{author}{J.~Bernhardt}, \bibinfo{author}{L.~Churilov},
\newblock \bibinfo{title}{Clustering clinical and health care processes using a
  novel measure of dissimilarity for variable-length sequences of ordinal
  states},
\newblock \bibinfo{journal}{Statistical methods in medical research}
  \bibinfo{volume}{29} (\bibinfo{year}{2020}) \bibinfo{pages}{3059--3075}.
\bibitem[{Bose and van~der Aalst(2009)}]{bose2009trace}
\bibinfo{author}{R.~J.~C. Bose}, \bibinfo{author}{W.~M. van~der Aalst},
\newblock \bibinfo{title}{Trace clustering based on conserved patterns: Towards
  achieving better process models},
\newblock in: \bibinfo{booktitle}{International Conference on Business Process
  Management}, \bibinfo{organization}{Springer}, \bibinfo{year}{2009}, pp.
  \bibinfo{pages}{170--181}.
\bibitem[{Prokofyeva et~al.(2019)Prokofyeva, Zaytsev, and
  Maltseva}]{prokofyeva2019application}
\bibinfo{author}{E.~S. Prokofyeva}, \bibinfo{author}{R.~D. Zaytsev},
  \bibinfo{author}{S.~V. Maltseva},
\newblock \bibinfo{title}{Application of modern data analysis methods to
  cluster the clinical pathways in urban medical facilities},
\newblock in: \bibinfo{booktitle}{2019 IEEE 21st Conference on Business
  Informatics (CBI)}, volume~\bibinfo{volume}{1}, \bibinfo{organization}{IEEE},
  \bibinfo{year}{2019}, pp. \bibinfo{pages}{75--83}.
\bibitem[{Chen et~al.(2019)Chen, Guo, Sun, and Lu}]{chen2019mining}
\bibinfo{author}{J.~Chen}, \bibinfo{author}{C.~Guo}, \bibinfo{author}{L.~Sun},
  \bibinfo{author}{M.~Lu},
\newblock \bibinfo{title}{Mining typical treatment duration patterns for
  rational drug use from electronic medical records},
\newblock \bibinfo{journal}{Journal of Systems Science and Systems Engineering}
  \bibinfo{volume}{28} (\bibinfo{year}{2019}) \bibinfo{pages}{602--620}.
\bibitem[{Aspland et~al.(2021)Aspland, Harper, Gartner, Webb, and
  Barrett-Lee}]{aspland2021modified}
\bibinfo{author}{E.~Aspland}, \bibinfo{author}{P.~R. Harper},
  \bibinfo{author}{D.~Gartner}, \bibinfo{author}{P.~Webb},
  \bibinfo{author}{P.~Barrett-Lee},
\newblock \bibinfo{title}{Modified {N}eedleman--{W}unsch algorithm for clinical
  pathway clustering},
\newblock \bibinfo{journal}{Journal of Biomedical Informatics}
  \bibinfo{volume}{115} (\bibinfo{year}{2021}) \bibinfo{pages}{103668}.
\bibitem[{Bean et~al.(2017)Bean, Stringer, Beeknoo, Teo, and
  Dobson}]{bean2017network}
\bibinfo{author}{D.~M. Bean}, \bibinfo{author}{C.~Stringer},
  \bibinfo{author}{N.~Beeknoo}, \bibinfo{author}{J.~Teo},
  \bibinfo{author}{R.~J. Dobson},
\newblock \bibinfo{title}{Network analysis of patient flow in two uk acute care
  hospitals identifies key sub-networks for {A}\&{E} performance},
\newblock \bibinfo{journal}{PloS one} \bibinfo{volume}{12}
  (\bibinfo{year}{2017}) \bibinfo{pages}{e0185912}.
\bibitem[{Hompes et~al.(2015)Hompes, Buijs, Van~der Aalst, Dixit, and
  Buurman}]{hompes2015discovering}
\bibinfo{author}{B.~Hompes}, \bibinfo{author}{J.~Buijs},
  \bibinfo{author}{W.~Van~der Aalst}, \bibinfo{author}{P.~Dixit},
  \bibinfo{author}{J.~Buurman},
\newblock \bibinfo{title}{Discovering deviating cases and process variants
  using trace clustering},
\newblock in: \bibinfo{booktitle}{Proceedings of the 27th Benelux Conference on
  Artificial Intelligence (BNAIC), November}, \bibinfo{year}{2015}, pp.
  \bibinfo{pages}{5--6}.
\bibitem[{Chambard et~al.(2021)Chambard, Guyet, NGuyen, and
  Audureau}]{Chambard_2021}
\bibinfo{author}{M.~Chambard}, \bibinfo{author}{T.~Guyet},
  \bibinfo{author}{Y.-L. NGuyen}, \bibinfo{author}{E.~Audureau},
\newblock \bibinfo{title}{Temporal phenotyping for characterisation of hospital
  care pathways of {COVID}19 patients},
\newblock in: \bibinfo{booktitle}{Advanced Analytics and Learning on Temporal
  Data}, \bibinfo{publisher}{Springer International Publishing},
  \bibinfo{year}{2021}, pp. \bibinfo{pages}{55--70}. \URLprefix
  \url{https://doi.org/10.1007%2F978-3-030-91445-5_4}.
  \DOIprefix\doi{10.1007/978-3-030-91445-5_4}.
\bibitem[{Kumar et~al.(2020)Kumar, Sampath, Imran, and
  Pradeep}]{Mohan_Kumar_2020}
\bibinfo{author}{K.~N.~M. Kumar}, \bibinfo{author}{S.~Sampath},
  \bibinfo{author}{M.~Imran}, \bibinfo{author}{N.~Pradeep},
\newblock \bibinfo{title}{Clustering diagnostic codes: Exploratory machine
  learning approach for preventive care of chronic diseases},
\newblock in: \bibinfo{booktitle}{Advances in Intelligent Systems and
  Computing}, \bibinfo{publisher}{Springer Singapore}, \bibinfo{year}{2020},
  pp. \bibinfo{pages}{551--564}. \URLprefix
  \url{https://doi.org/10.1007%2F978-981-15-5679-1_53}.
  \DOIprefix\doi{10.1007/978-981-15-5679-1_53}.
\bibitem[{Huang et~al.(2015)Huang, Lu, Iqbal, Lin, Nguyen, Yang, Wang, Li, Ma,
  Li, and Jian}]{Huang_2015}
\bibinfo{author}{C.-W. Huang}, \bibinfo{author}{R.~Lu},
  \bibinfo{author}{U.~Iqbal}, \bibinfo{author}{S.-H. Lin},
  \bibinfo{author}{P.~A. Nguyen}, \bibinfo{author}{H.-C. Yang},
  \bibinfo{author}{C.-F. Wang}, \bibinfo{author}{J.~Li}, \bibinfo{author}{K.-L.
  Ma}, \bibinfo{author}{Y.-C. Li}, \bibinfo{author}{W.-S. Jian},
\newblock \bibinfo{title}{A richly interactive exploratory data analysis and
  visualization tool using electronic medical records},
\newblock \bibinfo{journal}{{BMC} Medical Informatics and Decision Making}
  \bibinfo{volume}{15} (\bibinfo{year}{2015}). \URLprefix
  \url{https://doi.org/10.1186%2Fs12911-015-0218-7}.
  \DOIprefix\doi{10.1186/s12911-015-0218-7}.
\bibitem[{Steinhaeuser and Chawla(2009)}]{steinhaeuser2009network}
\bibinfo{author}{K.~Steinhaeuser}, \bibinfo{author}{N.~V. Chawla},
\newblock \bibinfo{title}{A network-based approach to understanding and
  predicting diseases},
\newblock in: \bibinfo{booktitle}{Social computing and behavioral modeling},
  \bibinfo{publisher}{Springer}, \bibinfo{year}{2009}, pp.
  \bibinfo{pages}{1--8}.
\bibitem[{Hanauer and Ramakrishnan(2013)}]{hanauer2013modeling}
\bibinfo{author}{D.~A. Hanauer}, \bibinfo{author}{N.~Ramakrishnan},
\newblock \bibinfo{title}{Modeling temporal relationships in large scale
  clinical associations},
\newblock \bibinfo{journal}{Journal of the American Medical Informatics
  Association} \bibinfo{volume}{20} (\bibinfo{year}{2013})
  \bibinfo{pages}{332--341}.
\bibitem[{Glicksberg et~al.(2016)Glicksberg, Li, Badgeley, Shameer, Kosoy,
  Beckmann, Pho, Hakenberg, Ma, Ayers, Hoffman, Dan~Li, Schadt, Patel, Chen,
  and Dudley}]{glicksberg2016comparative}
\bibinfo{author}{B.~S. Glicksberg}, \bibinfo{author}{L.~Li},
  \bibinfo{author}{M.~A. Badgeley}, \bibinfo{author}{K.~Shameer},
  \bibinfo{author}{R.~Kosoy}, \bibinfo{author}{N.~D. Beckmann},
  \bibinfo{author}{N.~Pho}, \bibinfo{author}{J.~Hakenberg},
  \bibinfo{author}{M.~Ma}, \bibinfo{author}{K.~L. Ayers},
  \bibinfo{author}{G.~E. Hoffman}, \bibinfo{author}{S.~Dan~Li},
  \bibinfo{author}{E.~E. Schadt}, \bibinfo{author}{C.~J. Patel},
  \bibinfo{author}{R.~Chen}, \bibinfo{author}{J.~T. Dudley},
\newblock \bibinfo{title}{{Comparative analyses of population-scale phenomic
  data in electronic medical records reveal race-specific disease networks}},
\newblock \bibinfo{journal}{Bioinformatics} \bibinfo{volume}{32}
  (\bibinfo{year}{2016}) \bibinfo{pages}{i101--i110}. \URLprefix
  \url{https://doi.org/10.1093/bioinformatics/btw282}.
  \DOIprefix\doi{10.1093/bioinformatics/btw282}.
\bibitem[{Kannan et~al.(2016)Kannan, Swartz, Kiani, Silberberg, Tsipras,
  Gomez-Cabrero, Alexanderson, and Tegn{\`e}r}]{kannan2016conditional}
\bibinfo{author}{V.~Kannan}, \bibinfo{author}{F.~Swartz},
  \bibinfo{author}{N.~A. Kiani}, \bibinfo{author}{G.~Silberberg},
  \bibinfo{author}{G.~Tsipras}, \bibinfo{author}{D.~Gomez-Cabrero},
  \bibinfo{author}{K.~Alexanderson}, \bibinfo{author}{J.~Tegn{\`e}r},
\newblock \bibinfo{title}{Conditional disease development extracted from
  longitudinal health care cohort data using layered network construction},
\newblock \bibinfo{journal}{Scientific reports} \bibinfo{volume}{6}
  (\bibinfo{year}{2016}) \bibinfo{pages}{1--14}.
\bibitem[{Dong et~al.(2021)Dong, Lee, Hertzberg, Simpson, and Ho}]{Dong_2021}
\bibinfo{author}{W.~Dong}, \bibinfo{author}{E.~W. Lee}, \bibinfo{author}{V.~S.
  Hertzberg}, \bibinfo{author}{R.~L. Simpson}, \bibinfo{author}{J.~C. Ho},
\newblock \bibinfo{title}{{GASP}: Graph-based approximate sequential pattern
  mining for electronic health records},
\newblock in: \bibinfo{booktitle}{New Trends in Database and Information
  Systems}, \bibinfo{publisher}{Springer International Publishing},
  \bibinfo{year}{2021}, pp. \bibinfo{pages}{50--60}. \URLprefix
  \url{https://doi.org/10.1007%2F978-3-030-85082-1_5}.
  \DOIprefix\doi{10.1007/978-3-030-85082-1_5}.
\bibitem[{Kushima et~al.(2018)Kushima, Honda, Le, Yamazaki, Araki, and
  Yokota}]{Kushima_2019}
\bibinfo{author}{M.~Kushima}, \bibinfo{author}{Y.~Honda},
  \bibinfo{author}{H.~H. Le}, \bibinfo{author}{T.~Yamazaki},
  \bibinfo{author}{K.~Araki}, \bibinfo{author}{H.~Yokota},
\newblock \bibinfo{title}{Extraction and graph structuring of variants by
  detecting common parts of frequent clinical pathways},
\newblock in: \bibinfo{booktitle}{International MultiConference of Engineers
  and Computer Scientists}, \bibinfo{year}{2018}, pp.
  \bibinfo{pages}{207--218}.
\bibitem[{Zhang et~al.(2017)Zhang, Liu, Li, and Cui}]{Zhang_2017}
\bibinfo{author}{S.~Zhang}, \bibinfo{author}{L.~Liu}, \bibinfo{author}{H.~Li},
  \bibinfo{author}{L.~Cui},
\newblock \bibinfo{title}{Collaborative prediction model of disease risk by
  mining electronic health records},
\newblock in: \bibinfo{booktitle}{Collaborate Computing: Networking,
  Applications and Worksharing}, \bibinfo{publisher}{Springer International
  Publishing}, \bibinfo{year}{2017}, pp. \bibinfo{pages}{71--82}. \URLprefix
  \url{https://doi.org/10.1007%2F978-3-319-59288-6_7}.
  \DOIprefix\doi{10.1007/978-3-319-59288-6_7}.
\bibitem[{fei Wang et~al.(2021)fei Wang, jing Wang, Peng, hao Ren, Gao, lun Li,
  Wang, feng Wang, jun Han, yu~Lyu, ming Huan, Chen, yan Wang, xin Shu, zhong
  Zhou, and Li}]{Wang_2021}
\bibinfo{author}{Y.~fei Wang}, \bibinfo{author}{J.~jing Wang},
  \bibinfo{author}{W.~Peng}, \bibinfo{author}{Y.~hao Ren},
  \bibinfo{author}{C.~Gao}, \bibinfo{author}{Y.~lun Li},
  \bibinfo{author}{R.~Wang}, \bibinfo{author}{X.~feng Wang},
  \bibinfo{author}{S.~jun Han}, \bibinfo{author}{J.~yu~Lyu},
  \bibinfo{author}{J.~ming Huan}, \bibinfo{author}{C.~Chen},
  \bibinfo{author}{H.~yan Wang}, \bibinfo{author}{Z.~xin Shu},
  \bibinfo{author}{X.~zhong Zhou}, \bibinfo{author}{W.~Li},
\newblock \bibinfo{title}{Identification of hypertension subgroups through
  topological analysis of symptom-based patient similarity},
\newblock \bibinfo{journal}{Chinese Journal of Integrative Medicine}
  \bibinfo{volume}{27} (\bibinfo{year}{2021}) \bibinfo{pages}{656--665}.
  \URLprefix \url{https://doi.org/10.1007/s11655-021-3336-3}.
  \DOIprefix\doi{10.1007/s11655-021-3336-3}.
\bibitem[{Maass and Kim(2020)}]{maass2020markov}
\bibinfo{author}{K.~Maass}, \bibinfo{author}{M.~Kim},
\newblock \bibinfo{title}{A markov decision process approach to optimizing
  cancer therapy using multiple modalities},
\newblock \bibinfo{journal}{Mathematical Medicine and Biology: a journal of the
  IMA} \bibinfo{volume}{37} (\bibinfo{year}{2020}) \bibinfo{pages}{22--39}.
\bibitem[{Huang et~al.(2018)Huang, Ge, Dong, He, and
  Duan}]{huang2018probabilistic}
\bibinfo{author}{Z.~Huang}, \bibinfo{author}{Z.~Ge}, \bibinfo{author}{W.~Dong},
  \bibinfo{author}{K.~He}, \bibinfo{author}{H.~Duan},
\newblock \bibinfo{title}{Probabilistic modeling personalized treatment
  pathways using electronic health records},
\newblock \bibinfo{journal}{Journal of biomedical informatics}
  \bibinfo{volume}{86} (\bibinfo{year}{2018}) \bibinfo{pages}{33--48}.
\bibitem[{Leontjeva et~al.(2016)Leontjeva, Conforti, Di~Francescomarino, Dumas,
  and Maggi}]{leontjeva2016complex}
\bibinfo{author}{A.~Leontjeva}, \bibinfo{author}{R.~Conforti},
  \bibinfo{author}{C.~Di~Francescomarino}, \bibinfo{author}{M.~Dumas},
  \bibinfo{author}{F.~M. Maggi},
\newblock \bibinfo{title}{Complex symbolic sequence encodings for predictive
  monitoring of business processes},
\newblock in: \bibinfo{booktitle}{International Conference on Business Process
  Management}, \bibinfo{organization}{Springer}, \bibinfo{year}{2016}, pp.
  \bibinfo{pages}{297--313}.
\bibitem[{Nagrecha et~al.(2017)Nagrecha, Thomas, Feldman, and
  Chawla}]{Nagrecha_2017}
\bibinfo{author}{S.~Nagrecha}, \bibinfo{author}{P.~B. Thomas},
  \bibinfo{author}{K.~Feldman}, \bibinfo{author}{N.~V. Chawla},
\newblock \bibinfo{title}{Predicting chronic heart failure using diagnoses
  graphs},
\newblock in: \bibinfo{booktitle}{Lecture Notes in Computer Science},
  \bibinfo{publisher}{Springer International Publishing}, \bibinfo{year}{2017},
  pp. \bibinfo{pages}{295--312}. \URLprefix
  \url{https://doi.org/10.1007%2F978-3-319-66808-6_20}.
  \DOIprefix\doi{10.1007/978-3-319-66808-6_20}.
\bibitem[{Bueno et~al.(2018)Bueno, Hommersom, Lucas, Lobo, and
  Rodrigues}]{Bueno_2018}
\bibinfo{author}{M.~L.~P. Bueno}, \bibinfo{author}{A.~Hommersom},
  \bibinfo{author}{P.~J.~F. Lucas}, \bibinfo{author}{M.~Lobo},
  \bibinfo{author}{P.~P. Rodrigues},
\newblock \bibinfo{title}{Modeling the dynamics of multiple disease occurrence
  by latent states},
\newblock in: \bibinfo{booktitle}{Lecture Notes in Computer Science},
  \bibinfo{publisher}{Springer International Publishing}, \bibinfo{year}{2018},
  pp. \bibinfo{pages}{93--107}. \URLprefix
  \url{https://doi.org/10.1007%2F978-3-030-00461-3_7}.
  \DOIprefix\doi{10.1007/978-3-030-00461-3_7}.
\bibitem[{Roder et~al.(2021)Roder, Zhao, Challam, Little, Elder, Kostadinovska,
  Woodland, and Currow}]{roder2021female}
\bibinfo{author}{D.~Roder}, \bibinfo{author}{G.~W. Zhao},
  \bibinfo{author}{S.~Challam}, \bibinfo{author}{A.~Little},
  \bibinfo{author}{E.~Elder}, \bibinfo{author}{G.~Kostadinovska},
  \bibinfo{author}{L.~Woodland}, \bibinfo{author}{D.~Currow},
\newblock \bibinfo{title}{Female breast cancer in {N}ew {S}outh {W}ales,
  {A}ustralia, by country of birth: implications for health-service delivery},
\newblock \bibinfo{journal}{BMC public health} \bibinfo{volume}{21}
  (\bibinfo{year}{2021}) \bibinfo{pages}{1--14}.
\bibitem[{Te~Marvelde et~al.(2019)Te~Marvelde, McNair, Whitfield, Autier,
  Boyle, Sullivan, and Thomas}]{te2019alignment}
\bibinfo{author}{L.~Te~Marvelde}, \bibinfo{author}{P.~McNair},
  \bibinfo{author}{K.~Whitfield}, \bibinfo{author}{P.~Autier},
  \bibinfo{author}{P.~Boyle}, \bibinfo{author}{R.~Sullivan},
  \bibinfo{author}{R.~J. Thomas},
\newblock \bibinfo{title}{Alignment with indices of a care pathway is
  associated with improved survival: An observational population-based study in
  colon cancer patients},
\newblock \bibinfo{journal}{EClinicalMedicine} \bibinfo{volume}{15}
  (\bibinfo{year}{2019}) \bibinfo{pages}{42--50}.
\bibitem[{Shahabi-Kargar et~al.(2020)Shahabi-Kargar, Johnston, Warner-Smith,
  Creighton, and Roder}]{shahabi2020differences}
\bibinfo{author}{Z.~Shahabi-Kargar}, \bibinfo{author}{A.~Johnston},
  \bibinfo{author}{M.~Warner-Smith}, \bibinfo{author}{N.~Creighton},
  \bibinfo{author}{D.~Roder},
\newblock \bibinfo{title}{Differences in breast cancer treatment pathways for
  women participating in screening through {B}reast{S}creen {N}ew {S}outh
  {W}ales ({BSNSW}).},
\newblock \bibinfo{journal}{Australasian Medical Journal} \bibinfo{volume}{13}
  (\bibinfo{year}{2020}).
\bibitem[{Li et~al.(2016)Li, Gupta, Rana, Luo, Venkatesh, Ashely, and
  Phung}]{Li_2016}
\bibinfo{author}{C.~Li}, \bibinfo{author}{S.~Gupta}, \bibinfo{author}{S.~Rana},
  \bibinfo{author}{W.~Luo}, \bibinfo{author}{S.~Venkatesh},
  \bibinfo{author}{D.~Ashely}, \bibinfo{author}{D.~Phung},
\newblock \bibinfo{title}{Toxicity prediction in cancer using multiple instance
  learning in a multi-task framework},
\newblock in: \bibinfo{booktitle}{Advances in Knowledge Discovery and Data
  Mining}, \bibinfo{publisher}{Springer International Publishing},
  \bibinfo{year}{2016}, pp. \bibinfo{pages}{152--164}. \URLprefix
  \url{https://doi.org/10.1007%2F978-3-319-31753-3_13}.
  \DOIprefix\doi{10.1007/978-3-319-31753-3_13}.
\bibitem[{Sun et~al.(2022)Sun, Douiri, and Gulliford}]{sun2022applying}
\bibinfo{author}{X.~Sun}, \bibinfo{author}{A.~Douiri},
  \bibinfo{author}{M.~Gulliford},
\newblock \bibinfo{title}{Applying machine learning algorithms to electronic
  health records to predict pneumonia after respiratory tract infection},
\newblock \bibinfo{journal}{Journal of Clinical Epidemiology}
  \bibinfo{volume}{145} (\bibinfo{year}{2022}) \bibinfo{pages}{154--163}.
\bibitem[{Kaur et~al.(2020)Kaur, Doja, and Ahmad}]{kaur2020time}
\bibinfo{author}{I.~Kaur}, \bibinfo{author}{M.~Doja},
  \bibinfo{author}{T.~Ahmad},
\newblock \bibinfo{title}{Time-range based sequential mining for survival
  prediction in prostate cancer},
\newblock \bibinfo{journal}{Journal of Biomedical Informatics}
  \bibinfo{volume}{110} (\bibinfo{year}{2020}) \bibinfo{pages}{103550}.
\bibitem[{Estiri et~al.(2020)Estiri, Vasey, and Murphy}]{Estiri_2020}
\bibinfo{author}{H.~Estiri}, \bibinfo{author}{S.~Vasey}, \bibinfo{author}{S.~N.
  Murphy},
\newblock \bibinfo{title}{Transitive sequential pattern mining for discrete
  clinical data},
\newblock in: \bibinfo{booktitle}{Artificial Intelligence in Medicine},
  \bibinfo{publisher}{Springer International Publishing}, \bibinfo{year}{2020},
  pp. \bibinfo{pages}{414--424}. \URLprefix
  \url{https://doi.org/10.1007%2F978-3-030-59137-3_37}.
  \DOIprefix\doi{10.1007/978-3-030-59137-3_37}.
\bibitem[{Vincent-Paulraj et~al.(2021)Vincent-Paulraj, Burnside, Coenen,
  Pirmohamed, and Walker}]{Vincent_Paulraj_2021}
\bibinfo{author}{A.~Vincent-Paulraj}, \bibinfo{author}{G.~Burnside},
  \bibinfo{author}{F.~Coenen}, \bibinfo{author}{M.~Pirmohamed},
  \bibinfo{author}{L.~Walker},
\newblock \bibinfo{title}{Sequential association rule mining revisited: A study
  directed at relational pattern mining for multi-morbidity},
\newblock in: \bibinfo{booktitle}{Lecture Notes in Computer Science},
  \bibinfo{publisher}{Springer International Publishing}, \bibinfo{year}{2021},
  pp. \bibinfo{pages}{241--253}. \URLprefix
  \url{https://doi.org/10.1007%2F978-3-030-91100-3_20}.
  \DOIprefix\doi{10.1007/978-3-030-91100-3_20}.
\bibitem[{Huang et~al.(2015)Huang, Dong, Ji, and Duan}]{Huang_2015a}
\bibinfo{author}{Z.~Huang}, \bibinfo{author}{W.~Dong}, \bibinfo{author}{L.~Ji},
  \bibinfo{author}{H.~Duan},
\newblock \bibinfo{title}{Outcome prediction in clinical treatment processes},
\newblock \bibinfo{journal}{Journal of Medical Systems} \bibinfo{volume}{40}
  (\bibinfo{year}{2015}). \URLprefix
  \url{https://doi.org/10.1007%2Fs10916-015-0380-6}.
  \DOIprefix\doi{10.1007/s10916-015-0380-6}.
\bibitem[{Boland et~al.(2015)Boland, Tatonetti, and Hripcsak}]{Boland_2015}
\bibinfo{author}{M.~R. Boland}, \bibinfo{author}{N.~P. Tatonetti},
  \bibinfo{author}{G.~Hripcsak},
\newblock \bibinfo{title}{Development and validation of a classification
  approach for extracting severity automatically from electronic health
  records},
\newblock \bibinfo{journal}{Journal of Biomedical Semantics}
  \bibinfo{volume}{6} (\bibinfo{year}{2015}). \URLprefix
  \url{https://doi.org/10.1186%2Fs13326-015-0010-8}.
  \DOIprefix\doi{10.1186/s13326-015-0010-8}.
\bibitem[{Maali et~al.(2018)Maali, Perez-Concha, Coiera, Roffe, Day, and
  Gallego}]{Maali_2018}
\bibinfo{author}{Y.~Maali}, \bibinfo{author}{O.~Perez-Concha},
  \bibinfo{author}{E.~Coiera}, \bibinfo{author}{D.~Roffe},
  \bibinfo{author}{R.~O. Day}, \bibinfo{author}{B.~Gallego},
\newblock \bibinfo{title}{Predicting 7-day, 30-day and 60-day all-cause
  unplanned readmission: a case study of a {S}ydney hospital},
\newblock \bibinfo{journal}{{BMC} Medical Informatics and Decision Making}
  \bibinfo{volume}{18} (\bibinfo{year}{2018}). \URLprefix
  \url{https://doi.org/10.1186%2Fs12911-017-0580-8}.
  \DOIprefix\doi{10.1186/s12911-017-0580-8}.
\bibitem[{{Wang} et~al.(2020){Wang}, {Chen}, {Chen}, and
  {Wang}}]{wang2020survivability}
\bibinfo{author}{K.-J. {Wang}}, \bibinfo{author}{J.-L. {Chen}},
  \bibinfo{author}{K.-H. {Chen}}, \bibinfo{author}{K.-M. {Wang}},
\newblock \bibinfo{title}{Survivability prognosis for lung cancer patients at
  different severity stages by a risk factor-based {B}ayesian network
  modeling},
\newblock \bibinfo{journal}{Journal of Medical Systems} \bibinfo{volume}{44}
  (\bibinfo{year}{2020}) \bibinfo{pages}{65}.
\bibitem[{Weiss and Page(2013)}]{Weiss_2013}
\bibinfo{author}{J.~C. Weiss}, \bibinfo{author}{D.~Page},
\newblock \bibinfo{title}{Forest-based point process for event prediction from
  electronic health records},
\newblock in: \bibinfo{booktitle}{Advanced Information Systems Engineering},
  \bibinfo{publisher}{Springer Berlin Heidelberg}, \bibinfo{year}{2013}, pp.
  \bibinfo{pages}{547--562}. \URLprefix
  \url{https://doi.org/10.1007%2F978-3-642-40994-3_35}.
  \DOIprefix\doi{10.1007/978-3-642-40994-3_35}.
\bibitem[{Du et~al.(2019)Du, Shi, Liu, and Liu}]{du2019variance}
\bibinfo{author}{G.~Du}, \bibinfo{author}{Y.~Shi}, \bibinfo{author}{A.~Liu},
  \bibinfo{author}{T.~Liu},
\newblock \bibinfo{title}{Variance risk identification and treatment of
  clinical pathway by integrated {B}ayesian network and association rules
  mining},
\newblock \bibinfo{journal}{Entropy} \bibinfo{volume}{21}
  (\bibinfo{year}{2019}) \bibinfo{pages}{1191}.
\bibitem[{Wang et~al.(2012)Wang, Lee, Hu, Sun, Ebadollahi, and
  Laine}]{wang2012framework}
\bibinfo{author}{F.~Wang}, \bibinfo{author}{N.~Lee}, \bibinfo{author}{J.~Hu},
  \bibinfo{author}{J.~Sun}, \bibinfo{author}{S.~Ebadollahi},
  \bibinfo{author}{A.~F. Laine},
\newblock \bibinfo{title}{A framework for mining signatures from event
  sequences and its applications in healthcare data},
\newblock \bibinfo{journal}{IEEE transactions on pattern analysis and machine
  intelligence} \bibinfo{volume}{35} (\bibinfo{year}{2012})
  \bibinfo{pages}{272--285}.
\bibitem[{Nguyen et~al.(2015)Nguyen, Luo, Phung, and Venkatesh}]{Nguyen_2015}
\bibinfo{author}{D.~Nguyen}, \bibinfo{author}{W.~Luo},
  \bibinfo{author}{D.~Phung}, \bibinfo{author}{S.~Venkatesh},
\newblock \bibinfo{title}{Understanding toxicities and complications of cancer
  treatment: A data mining approach},
\newblock in: \bibinfo{booktitle}{{AI} 2015: Advances in Artificial
  Intelligence}, \bibinfo{publisher}{Springer International Publishing},
  \bibinfo{year}{2015}, pp. \bibinfo{pages}{431--443}. \URLprefix
  \url{https://doi.org/10.1007%2F978-3-319-26350-2_38}.
  \DOIprefix\doi{10.1007/978-3-319-26350-2_38}.
\bibitem[{Chen et~al.(2020)Chen, Sun, Guo, and Xie}]{chen2020fusion}
\bibinfo{author}{J.~Chen}, \bibinfo{author}{L.~Sun}, \bibinfo{author}{C.~Guo},
  \bibinfo{author}{Y.~Xie},
\newblock \bibinfo{title}{A fusion framework to extract typical treatment
  patterns from electronic medical records},
\newblock \bibinfo{journal}{Artificial intelligence in medicine}
  \bibinfo{volume}{103} (\bibinfo{year}{2020}) \bibinfo{pages}{101782}.
\bibitem[{Liu et~al.(2020)Liu, Zhang, Di, and Chen}]{liu2020elmv}
\bibinfo{author}{L.~J. Liu}, \bibinfo{author}{H.~Zhang},
  \bibinfo{author}{J.~Di}, \bibinfo{author}{J.~Chen},
\newblock \bibinfo{title}{{ELMV}: an ensemble-learning approach for analyzing
  electrical health records with significant missing values},
\newblock in: \bibinfo{booktitle}{Proceedings of the 11th ACM International
  Conference on Bioinformatics, Computational Biology and Health Informatics},
  \bibinfo{year}{2020}, pp. \bibinfo{pages}{1--10}.
\bibitem[{Xue et~al.(2019)Xue, Klabjan, and Luo}]{xue2019mixture}
\bibinfo{author}{Y.~Xue}, \bibinfo{author}{D.~Klabjan},
  \bibinfo{author}{Y.~Luo},
\newblock \bibinfo{title}{Mixture-based multiple imputation model for clinical
  data with a temporal dimension},
\newblock in: \bibinfo{booktitle}{2019 IEEE International Conference on Big
  Data (Big Data)}, \bibinfo{organization}{IEEE}, \bibinfo{year}{2019}, pp.
  \bibinfo{pages}{245--252}.
\bibitem[{Huda et~al.(2016)Huda, Yearwood, Jelinek, Hassan, Fortino, and
  Buckland}]{huda2016hybrid}
\bibinfo{author}{S.~Huda}, \bibinfo{author}{J.~Yearwood},
  \bibinfo{author}{H.~F. Jelinek}, \bibinfo{author}{M.~M. Hassan},
  \bibinfo{author}{G.~Fortino}, \bibinfo{author}{M.~Buckland},
\newblock \bibinfo{title}{A hybrid feature selection with ensemble
  classification for imbalanced healthcare data: A case study for brain tumor
  diagnosis},
\newblock \bibinfo{journal}{IEEE access} \bibinfo{volume}{4}
  (\bibinfo{year}{2016}) \bibinfo{pages}{9145--9154}.
\bibitem[{Ray and Wimalasiri(2006)}]{Ray_2006}
\bibinfo{author}{P.~Ray}, \bibinfo{author}{J.~Wimalasiri},
\newblock \bibinfo{title}{The need for technical solutions for maintaining the
  privacy of {EHR}},
\newblock in: \bibinfo{booktitle}{2006 International Conference of the IEEE
  Engineering in Medicine and Biology Society}, \bibinfo{year}{2006}, pp.
  \bibinfo{pages}{4686--4689}. \DOIprefix\doi{10.1109/IEMBS.2006.260862}.
\bibitem[{Myers and Stevens(2016)}]{Myers2016}
\bibinfo{author}{L.~Myers}, \bibinfo{author}{J.~Stevens}, \bibinfo{title}{Using
  {EHR} to Conduct Outcome and Health Services Research},
  \bibinfo{publisher}{Springer International Publishing},
  \bibinfo{address}{Cham}, \bibinfo{year}{2016}, pp. \bibinfo{pages}{61--70}.
  \URLprefix \url{https://doi.org/10.1007/978-3-319-43742-2_7}.
  \DOIprefix\doi{10.1007/978-3-319-43742-2_7}.
\bibitem[{{World Health Organization}(2004)}]{ICD_10}
\bibinfo{author}{{World Health Organization}}, \bibinfo{title}{Icd-10 :
  international statistical classification of diseases and related health
  problems : tenth revision}, \bibinfo{year}{2004}.
\bibitem[{{Healthcare Cost and Utilization Project}(2019)}]{CCS}
\bibinfo{author}{{Healthcare Cost and Utilization Project}},
  \bibinfo{title}{{Clinical Classifications Software (CCS) for ICD-10-PCS (beta
  version)}},
  \bibinfo{howpublished}{\url{https://www.hcup-us.ahrq.gov/toolssoftware/ccs10/ccs10.jsp}},
  \bibinfo{year}{2019}. \bibinfo{note}{Accessed: 2022-03-11}.
\bibitem[{Johnson et~al.(2016)Johnson, Pollard, Shen, Li-Wei, Feng, Ghassemi,
  Moody, Szolovits, Celi, and Mark}]{johnson2016mimic}
\bibinfo{author}{A.~E. Johnson}, \bibinfo{author}{T.~J. Pollard},
  \bibinfo{author}{L.~Shen}, \bibinfo{author}{H.~L. Li-Wei},
  \bibinfo{author}{M.~Feng}, \bibinfo{author}{M.~Ghassemi},
  \bibinfo{author}{B.~Moody}, \bibinfo{author}{P.~Szolovits},
  \bibinfo{author}{L.~A. Celi}, \bibinfo{author}{R.~G. Mark},
\newblock \bibinfo{title}{{MIMIC-III}, a freely accessible critical care
  database},
\newblock \bibinfo{journal}{Scientific data} \bibinfo{volume}{3}
  (\bibinfo{year}{2016}) \bibinfo{pages}{1--9}.
\bibitem[{Johnson et~al.(2020)Johnson, Bulgarelli, Pollard, Horng, Celi, and
  Mark~IV}]{johnson2020mimic}
\bibinfo{author}{A.~Johnson}, \bibinfo{author}{L.~Bulgarelli},
  \bibinfo{author}{T.~Pollard}, \bibinfo{author}{S.~Horng},
  \bibinfo{author}{L.~A. Celi}, \bibinfo{author}{R.~Mark~IV},
\newblock \bibinfo{title}{{MIMIC-IV} (version 0.4)},
\newblock \bibinfo{journal}{PhysioNet}  (\bibinfo{year}{2020}).
\bibitem[{van Dongen(2011)}]{vanDongen2011Dutch}
\bibinfo{author}{B.~van Dongen},
\newblock \bibinfo{title}{Real-life event logs - hospital log},
\newblock \bibinfo{journal}{4TU.ResearchData.Dataset}  (\bibinfo{year}{2011}).
  \DOIprefix\doi{10.4121/uuid:d9769f3d-0ab0-4fb8-803b-0d1120ffcf54}.
\bibitem[{Raghunathan(2021)}]{Raghunathan_2021}
\bibinfo{author}{T.~E. Raghunathan},
\newblock \bibinfo{title}{Synthetic data},
\newblock \bibinfo{journal}{Annual Review of Statistics and Its Application}
  \bibinfo{volume}{8} (\bibinfo{year}{2021}) \bibinfo{pages}{129--140}.
  \URLprefix \url{https://doi.org/10.1146%2Fannurev-statistics-040720-031848}.
  \DOIprefix\doi{10.1146/annurev-statistics-040720-031848}.
\bibitem[{Goncalves et~al.(2020)Goncalves, Ray, Soper, Stevens, Coyle, and
  Sales}]{Goncalves_2020}
\bibinfo{author}{A.~Goncalves}, \bibinfo{author}{P.~Ray},
  \bibinfo{author}{B.~Soper}, \bibinfo{author}{J.~Stevens},
  \bibinfo{author}{L.~Coyle}, \bibinfo{author}{A.~P. Sales},
\newblock \bibinfo{title}{Generation and evaluation of synthetic patient data},
\newblock \bibinfo{journal}{{BMC} Medical Research Methodology}
  \bibinfo{volume}{20} (\bibinfo{year}{2020}). \URLprefix
  \url{https://doi.org/10.1186%2Fs12874-020-00977-1}.
  \DOIprefix\doi{10.1186/s12874-020-00977-1}.
\bibitem[{Tucker et~al.(2020)Tucker, Wang, Rotalinti, and Myles}]{Tucker_2020}
\bibinfo{author}{A.~Tucker}, \bibinfo{author}{Z.~Wang},
  \bibinfo{author}{Y.~Rotalinti}, \bibinfo{author}{P.~Myles},
\newblock \bibinfo{title}{Generating high-fidelity synthetic patient data for
  assessing machine learning healthcare software},
\newblock \bibinfo{journal}{NPJ Digital Medicine} \bibinfo{volume}{3}
  (\bibinfo{year}{2020}). \URLprefix
  \url{https://doi.org/10.1038%2Fs41746-020-00353-9}.
  \DOIprefix\doi{10.1038/s41746-020-00353-9}.
\bibitem[{Gashler and Martinez(2011)}]{gashler2011temporal}
\bibinfo{author}{M.~Gashler}, \bibinfo{author}{T.~Martinez},
\newblock \bibinfo{title}{Temporal nonlinear dimensionality reduction},
\newblock in: \bibinfo{booktitle}{The 2011 International Joint Conference on
  Neural Networks}, \bibinfo{organization}{IEEE}, \bibinfo{year}{2011}, pp.
  \bibinfo{pages}{1959--1966}.
\bibitem[{Ali et~al.(2019)Ali, Jones, Xie, and Williams}]{ali2019timecluster}
\bibinfo{author}{M.~Ali}, \bibinfo{author}{M.~W. Jones},
  \bibinfo{author}{X.~Xie}, \bibinfo{author}{M.~Williams},
\newblock \bibinfo{title}{Timecluster: dimension reduction applied to temporal
  data for visual analytics},
\newblock \bibinfo{journal}{The Visual Computer} \bibinfo{volume}{35}
  (\bibinfo{year}{2019}) \bibinfo{pages}{1013--1026}.
\bibitem[{Lewandowski et~al.(2010)Lewandowski, Martinez-del Rincon, Makris, and
  Nebel}]{lewandowski2010temporal}
\bibinfo{author}{M.~Lewandowski}, \bibinfo{author}{J.~Martinez-del Rincon},
  \bibinfo{author}{D.~Makris}, \bibinfo{author}{J.-C. Nebel},
\newblock \bibinfo{title}{Temporal extension of {L}aplacian eigenmaps for
  unsupervised dimensionality reduction of time series},
\newblock in: \bibinfo{booktitle}{2010 20th International Conference on Pattern
  Recognition}, \bibinfo{organization}{IEEE}, \bibinfo{year}{2010}, pp.
  \bibinfo{pages}{161--164}.
\bibitem[{Liu and Qin(2022)}]{liu2022interpretable}
\bibinfo{author}{Y.~Liu}, \bibinfo{author}{S.~Qin},
\newblock \bibinfo{title}{An interpretable machine learning approach for
  predicting hospital length of stay and readmission},
\newblock in: \bibinfo{booktitle}{International Conference on Advanced Data
  Mining and Applications}, \bibinfo{organization}{Springer},
  \bibinfo{year}{2022}, pp. \bibinfo{pages}{73--85}.
\bibitem[{Lipton(2018)}]{lipton2018mythos}
\bibinfo{author}{Z.~C. Lipton},
\newblock \bibinfo{title}{The mythos of model interpretability: In machine
  learning, the concept of interpretability is both important and slippery.},
\newblock \bibinfo{journal}{Queue} \bibinfo{volume}{16} (\bibinfo{year}{2018})
  \bibinfo{pages}{31--57}.
\bibitem[{Wilkinson et~al.(2016)Wilkinson, Dumontier, Aalbersberg, Appleton,
  Axton, Baak, Blomberg, Boiten, da~Silva~Santos, Bourne
  et~al.}]{wilkinson2016fair}
\bibinfo{author}{M.~D. Wilkinson}, \bibinfo{author}{M.~Dumontier},
  \bibinfo{author}{I.~J. Aalbersberg}, \bibinfo{author}{G.~Appleton},
  \bibinfo{author}{M.~Axton}, \bibinfo{author}{A.~Baak},
  \bibinfo{author}{N.~Blomberg}, \bibinfo{author}{J.-W. Boiten},
  \bibinfo{author}{L.~B. da~Silva~Santos}, \bibinfo{author}{P.~E. Bourne},
  et~al.,
\newblock \bibinfo{title}{The fair guiding principles for scientific data
  management and stewardship},
\newblock \bibinfo{journal}{Scientific data} \bibinfo{volume}{3}
  (\bibinfo{year}{2016}) \bibinfo{pages}{1--9}.

\end{thebibliography}

%

\end{document}